\pdfoutput=1

\documentclass[11pt]{article}

\usepackage[]{acl}

\usepackage{times}
\usepackage{latexsym}

\usepackage[T1]{fontenc}

\usepackage[utf8]{inputenc}

\usepackage{microtype}

\usepackage{algorithm}
\usepackage{algorithmic}

\usepackage{multirow} %
\usepackage{makecell}
\usepackage{graphicx}
\usepackage{amsfonts}
\usepackage{eucal}
\usepackage{tabularx}
\usepackage{xcolor, soul}
\usepackage{caption}
\usepackage{amsmath}%
\usepackage{MnSymbol}%
\usepackage{wasysym}%
\usepackage{lipsum,subcaption}

\title{Stylized Knowledge-Grounded Dialogue Generation \\ via Disentangled Template Rewriting}

\author{Qingfeng Sun \quad Can Xu \quad {\bf Huang Hu} \quad Yujing Wang \quad {\bf Jian Miao} \\ \quad {\bf Xiubo Geng} \quad {\bf Yining Chen} \quad {\bf Fei Xu}  \quad {\bf Daxin Jiang}\thanks{\quad Corresponding author.}  \\
      Microsoft, Beijing, China\\
      \texttt{\{qins,caxu,huahu,yujwang,jianm,}\\
      \texttt{xigeng,yinichen,fexu,djiang\}@microsoft.com}}
      
\begin{document}
\maketitle

\begin{abstract}
Current Knowledge-Grounded Dialogue Generation (KDG) models specialize in producing rational and factual responses. However, to establish long-term relationships with users, the KDG model needs the capability to generate responses in a desired style or sentiment. Thus, we study a new problem: Stylized Knowledge-Grounded Dialogue Generation (SKDG). It presents two challenges: (1) How to train a SKDG model where no $<$context, knowledge, stylized response$>$ triples are available. (2) How to cohere with context and preserve the knowledge when generating a stylized response. In this paper, we propose a novel disentangled template rewriting (DTR) method which generates responses via combing disentangled style templates (from monolingual stylized corpus) and content templates (from KDG corpus). The entire framework is end-to-end differentiable and learned without supervision. Extensive experiments on two benchmarks indicate that DTR achieves a significant improvement on all evaluation metrics compared with previous state-of-the-art stylized dialogue generation methods. Besides, DTR achieves comparable performance with the state-of-the-art KDG methods in standard KDG evaluation setting.




\end{abstract}

\section{Introduction}
\begin{figure}[ht]
\centering
     \includegraphics[width=0.48\textwidth, scale=1, trim=335 140 200 195,clip]{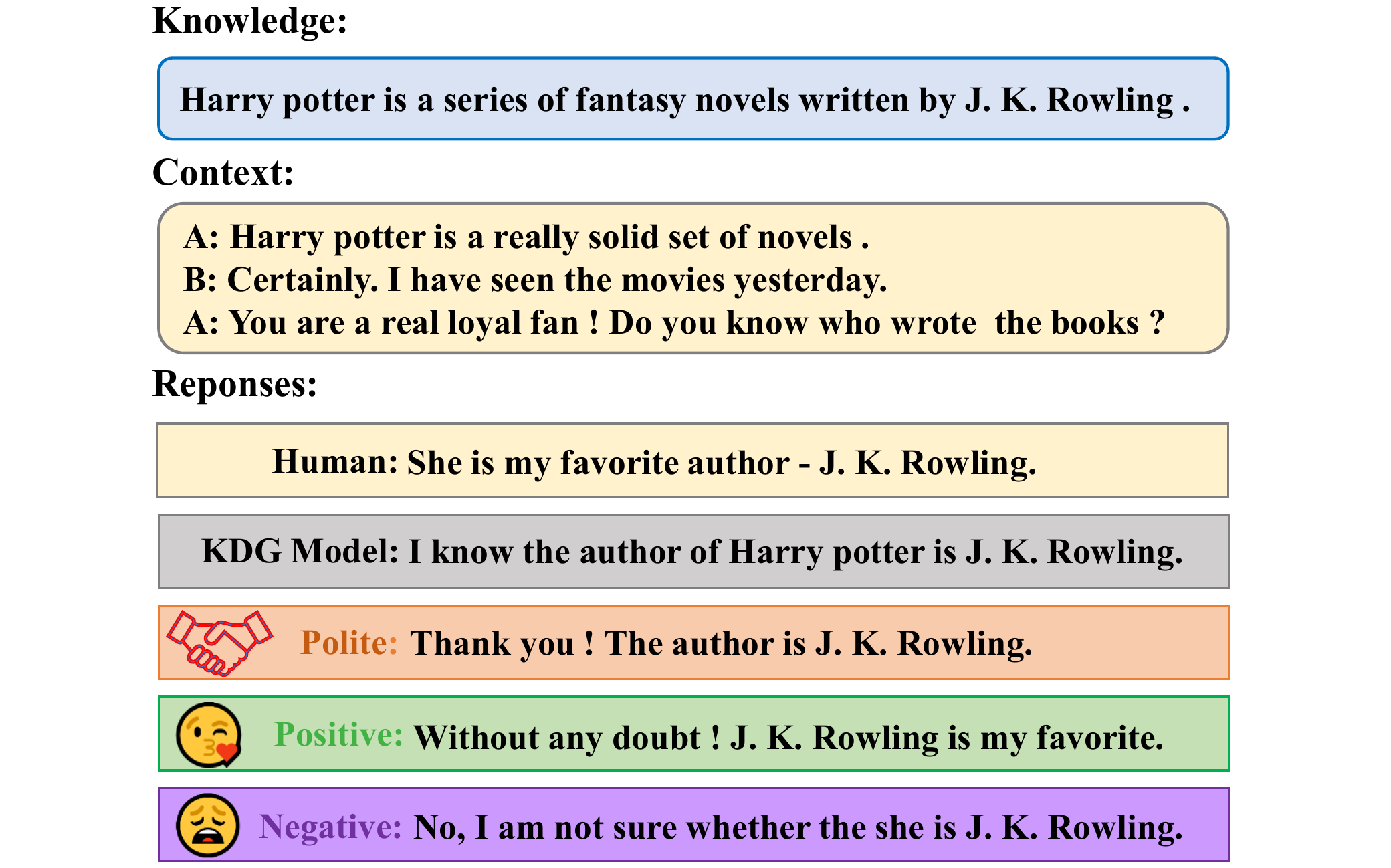}
     \caption{The KDG models only produce a pedantic response, which lacks emotion and attraction compared with the responses with polite style, positive and negative sentiments.}
     \label{fig:intro_example}
\end{figure}
Every good conversational agent needs the ability to generate good responses, which are not only knowledgeable and coherent with contexts but also have abundant and desirable styles and sentiments \cite{rashkin2018towards,smith2020controlling,zhou2020design}. Such an agent can deliver depth dialogues on various topics and yield more engaging and vivacious conversations to attract more users. In other words, rational and perceptual thought are all necessary for a perfect dialogue agent. Nevertheless, most existing Knowledge-Grounded Dialogue Generation (KDG) methods \cite{Dinan2019kgc,kim2020sequential,zhao2020knowledge} pay more attention to the former and ignore the latter. Specifically, let's claim our motivation: The previous KDG works mainly focus on selecting knowledge and expressing knowledge in response accurately. However, the excessive emphasis on knowledge makes the KDG models tend to mechanically copy large sections from the unstructured knowledge (e.g., Wikipedia). As a result, the responses from the KDG models reflect a ``pedantic" style (i.e., use very technical terms and language), making the conversation less engaging and less natural.

In this paper, we are aiming to have the first attempt to incorporate stylized-text-generation into KDG to tackle the above challenge. As shown in Figure \ref{fig:intro_example}, the KDG model takes the context and related document as input and outputs a knowledgeable but pedantic response corresponding to the polite one,  which makes people feel respected and comfortable. In the meanwhile, the polite, positive responses all show bright and lively styles which not only are able to condense the core meaning of the response, but also sound appealing to the users for more exposure and memorableness.

\begin{figure*}[bht]
\centering
\includegraphics[scale=2, width=0.97\textwidth, trim=10 175 10 162,clip]{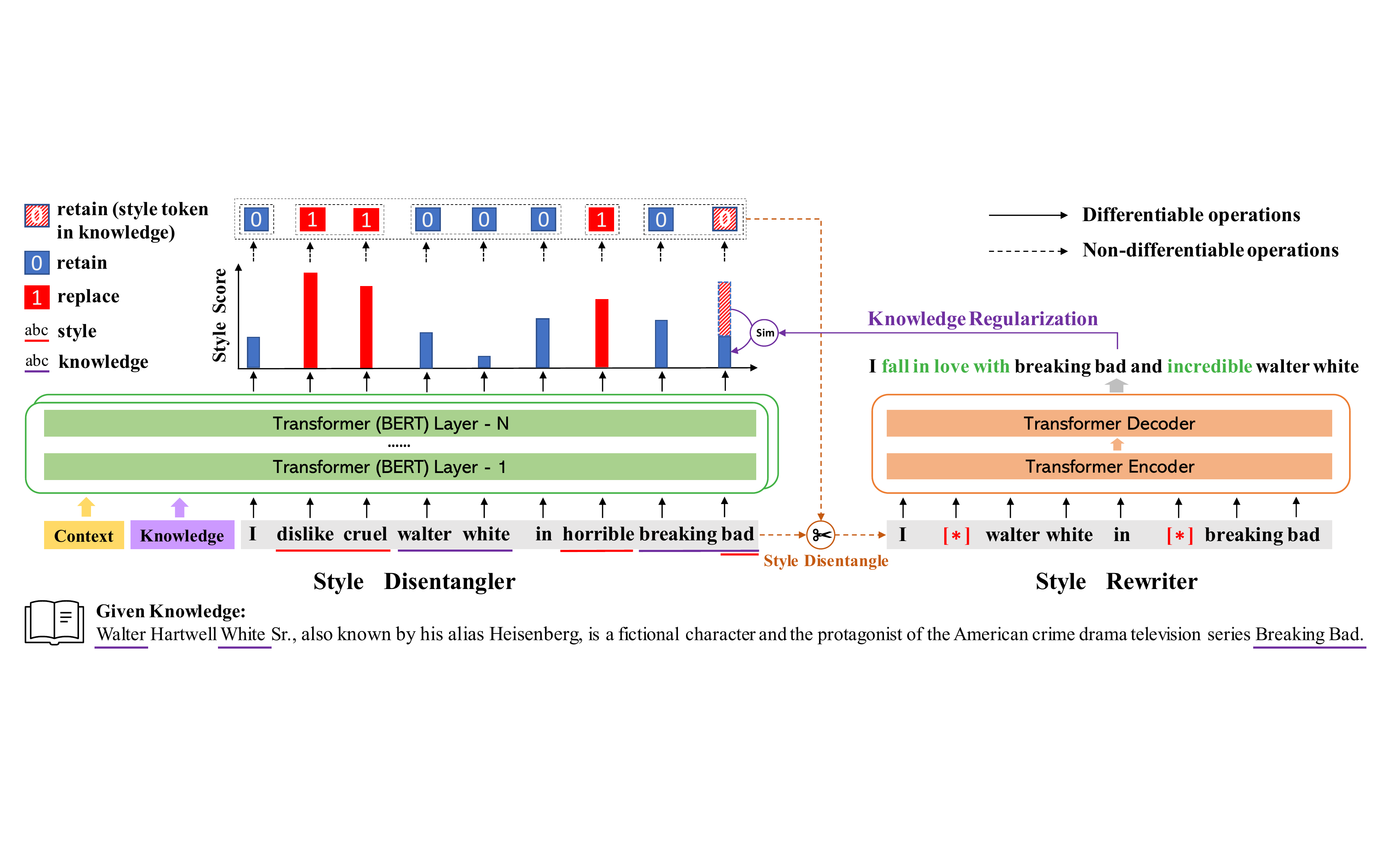}
\caption{Overview of DTR. The sequential style disentangler can find out and replace the style-related tokens in generated response with [*] and produce a template. Then, the style rewriter transfers the template to a new response in the target style.}
\label{fig:my_model}
\end{figure*}

\begin{figure}[ht]
\centering
     \includegraphics[width=0.5\textwidth, scale=1, trim=165 295 303 176,clip]{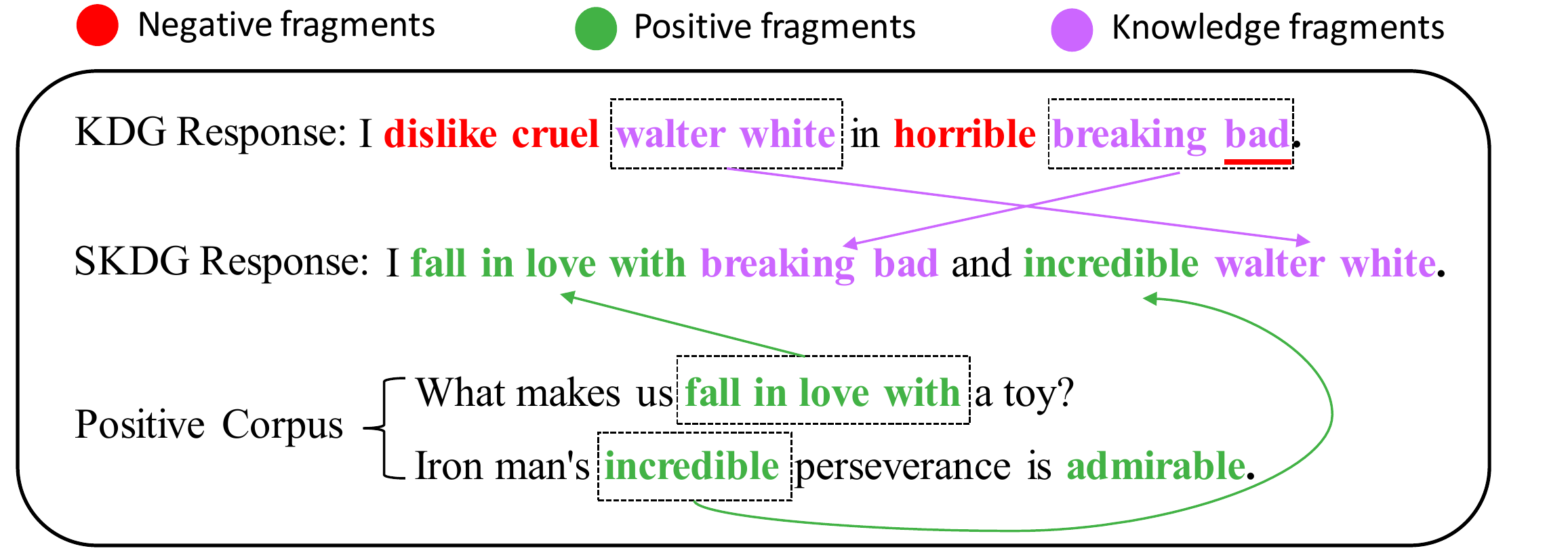}
     \caption{Our model combines the style-related fragments from the style corpus and the knowledge fragments from the KDG response to generate a response in a desired sentiment or style.}
     \label{fig:intro_text_case}
\end{figure}

Specifically, we formulate a new problem: Stylized Knowledge-Grounded Dialogue Generation (SKDG). That is, the responses provided by a model should be coherent with the dialogue contexts and be consistent with the given knowledge and a designated style or sentiment.
The challenges lie in two aspects: (1) As lacking stylized knowledge-grounded dialogue triples (i.e., $<$context, knowledge, stylized response$>$), we need to train the SKDG model jointly by both independent knowledge-grounded dialogues and monolingual corpus with a target style  or sentiment. (2) In addition to being coherent with context and consistent with target style / sentiment, a good response from SKDG needs to ensure objective correctness in the knowledge section. Especially when the given knowledge contains style-related content, existing stylized dialogue generation (SDG) models  \cite{zheng2020stylized,yang2020styledgpt} may undermine the correctness of knowledge section. 
For example, in case of negative-to-positive sentiment transfer shown in Figure \ref{fig:intro_text_case}, the first two negative fragments of KDG response - `` dislike cruel'' and ``horrible'' should be modified to positive fragments, but the third `` bad '' should be retained to maintain the original meaning of knowledge section. 




Hence, our motivation is: on the one hand, bridging the separate knowledge-grounded response generation and stylized rewriting by sharing a disentangled template (addressing challenge (1)); on the other hand, enhancing the fidelity regarding to given knowledge by using a reinforcement learning approach (addressing challenge (2)).

To achieve this goal, we propose a new paradigm: Generate-Disentangle-Rewrite. Firstly, given a dialogue context and the associated external knowledge, a KDG model is adopted to generate a response. Then as shown in Figure \ref{fig:my_model} and \ref{fig:intro_text_case}, we leverage a sequential style disentangler to delete style-related fragments from the KDG response to form a style-agnostic template. Then the rewriter rewrites the entire template token-by-token, injecting style-related fragments in the process, to generate a vivid and informative response in the desired style. As there is no supervision on the style disentangler and the style rewriter, we propose a reinforcement learning-based method to train the style disentangling and style rewriting in an end-to-end manner using a style intensity reward and a semantic similarity reward. The huge joint action space of the two modules fragile the training, thus we propose a novel weakly supervised stylistic template disentangle method to initialize both the disentangler and the rewriter. As a result, our method successfully produces the knowledgeable response in the desired style without any paired training data. 


We name our model \textbf{DTR} standing for ``\textbf{D}isentangled  \textbf{T}emplate  \textbf{R}ewriting''. We demonstrate this approach using knowledge-grounded dialogues from Wizard of Wikipedia \citep{Dinan2019kgc} and Topical Chat \citep{Gopalakrishnan2019} with three sets of sentences with distinct sentiments  (positive, negative)  and styles  (polite). Automatic and human evaluations show that our method significantly outperforms competitive baselines with a large margin in generating coherent and knowledgeable dialogue responses while rendering stronger stylistic features.

Our contributions are three-fold: (1) To the best of our knowledge, it is the first work on the generation of stylized knowledge-grounded responses without any labeled paired data for style-specific context-knowledge-response. (2) We proposed a stylized knowledge-grounded dialogue generation method via disentangled template rewriting. To optimize the model, we propose a reinforcement learning approach with a novel weakly supervised method to guide the learning of both disentangler and rewriter.
(3) Extensive experiments on two benchmarks indicate that DTR significantly outperforms previous state-of-the-art SDG methods on all evaluation metrics. Besides, DTR achieves comparable performance with the state-of-the-art KDG methods in the standard KDG evaluation setting. Our source code will be released at~\url{https://github.com/victorsungo/SKDG-DTR}. 
\section{Related Work}
\noindent\textbf{Knowledge-Grounded Dialogue Generation} has attracted broad interest in recent years, where the knowledge could be obtained from documents \cite{Dinan2019kgc, kim2020sequential,rashkin-etal-2021-increasing} and images \cite{shuster2018image,yang2020open,liang2021maria}. Our study considers document-grounded dialogue generation. With the  rapid development of pre-training techniques, \citet{zhao2020low-resource} proposes pre-trained disentangled decoder, and \citet{li2020zero} proves that the KDG models could achieve comparable performance with state-of-the-art supervised methods through an unsupervised learning method. Rather than testing new architectures on the benchmarks, our main contribution lies in the investigation of transferring the pedantic and factual knowledge-grounded responses into a desired  style or sentiment, which roots in the requirement from practice.

\noindent\textbf{Text Style and Sentiment Transfer} was inspired by visual style transfer \citep{imagestyletransfer2016, CycleGAN2017}, and many methods have made remarkable work in text style transfer, which aims to alter the style attributes of text while preserving the content. A prevalent idea of  style transfer is to disentangle the content and style of text \cite{AAAI1817015, li-etal-2018-delete,jin-etal-2020-hooks,wen-etal-2020-decode,zhu-etal-2021-neural} or leverage the back-translation \citep{lample2019multiple-attribute, li2021stylized}. Stylized dialogue generation has attracted numerous attention in recent years \citep{niu2018polite, gao2019stylefusion}. Different from style transfer, stylized dialogue generation requires that the response is also coherent with its context. 

\noindent\textbf{Stylized Dialogue Generation} refers to generate a dialogue response in the target style.   \citet{akama-etal-2017-generating} first train a response generation model on a dialog corpus then use a style corpus to fine-tune the model. \citet{yang-etal-2020-styledgpt} builds a pre-trained language model and devise both a word-level loss and a sentence-level loss to fine-tune the pre-trained model towards the target style. \citet{https://doi.org/10.48550/arxiv.2004.02202} proposes an information guided reinforcement learning strategy to better balance the trade-off between the stylistic expression and the content quality. \citet{https://doi.org/10.48550/arxiv.2110.08515} blends textual and visual responses to make the dialogue style more attractive and vivid. \citet{zheng2020stylized} captures stylistic features embedded in unpaired texts, \citet{su2020prototypetostyle} uses the pointwise mutual information (PMI) to determine stylistic word, \citet{yang2020styledgpt} adopts pre-trained models to tackle the open-domain stylized response generation.

We propose a novel disentangled template rewriting approach as the first attempt to study stylized knowledge-grounded dialogue generation without any supervised style-specific context-knowledge-response triples data.

\section{Task Definition}


For the SKDG task, our model is trained on a dialogue dataset $\mathcal{D}_{c}= \{(K_{i}, U_{i}, Y_{i}) \}^{N}_{i=1}$ and a style corpus $\mathcal{D}_{s} = \{ T_{i}\}^{M}_{i=1} $, where  $\forall{(K_{i}, U_{i}, Y_{i})} \in \mathcal{D}_{c}$, $U_{i}$ is a dialogue context, $K_{i}$ a external document that contains relevant knowledge regarding to $U_{i}$ and $Y_{i}$ a response to $U_{i}$, and $\forall{T_{i}} \in \mathcal{D}_{s}$, $T_{i}$ is a piece of text in the target style $\mathcal{S}$. We don't assume that there exists triples $\{(K, U, Y^{'})\}$ with $Y{'}$ expressed in the style or sentiment $\mathcal{S}$, e.g., $\mathcal{S} = \{``polite",``positive", ``negative"\}$. Our goal is to learn a generation method $P(Y |K,U,S)$ with $\mathcal{D}_{c}$ and $\mathcal{D}_{s}$, thus given a  document $K$ and a context $U$, one can generate a response $Y$ following the desired style $\mathcal{S}$, where $Y$ also coheres with context and preserves the knowledge.


\section{Approach}

Heading for learning an effective disentangled template rewriting model for SKDG task, we need to deal with several challenges: (1) how to distinguish the style-related fragments from a given sentence without any supervision;(2) how to retain the style-related fragments in knowledge section to defend the completeness;(3) how to rewrite the disentangled template holistically instead of inserting a few style words, thus to enhance fluency and diversity.

Our DTR model is made up of a knowledge-grounded response generator $\mathcal{G}_G$, a sequential style disentangler $\mathcal{F}$ and a style rewriter $\mathcal{G}_R$. Given a dialogue context $U$ and its associated knowledge $K$, we first use $\mathcal{G}_G$ to generate a response $\overline{Y}$. Figure \ref{fig:my_model} illustrates the cooperation of $\mathcal{F}$ and $\mathcal{G}_R$. The former reads $\overline{Y}$ and disentangles the style-related content from $\overline{Y}$ to form a style-agnostic template sequence $\widetilde{Y}$, which is further provided as input to $\mathcal{G}_R$ to generate the transferred response $\hat{Y}$ in a target style. Since $\widetilde{Y}$ is discrete, the major hinder of learning $\mathcal{F}$ lies in the gradient is not differentiable. To cope with the challenge, we exploit a reinforcement learning approach to optimize $\mathcal{F}$ leveraging the signals from $\mathcal{G}_R$. 

So why do we need a Disentangler + Rewriter architecture? The previous SDG methods fuse the knowledge and style into mixed representation and decode a response. Due to the difficulty of mixing knowledge and style implicitly under the unsupervised setting, it is possible to lose knowledge or style in the decoding stage. Motivated by this, we propose to decouple the response generation into two relatively independent processes: 1) knowledge fragments generation 2) style fragments generation. The knowledge fragments and style fragments are explicitly composited into the response in the final stage. Such a method ensure the knowledge is successfully presented in the final output. The disentangler plays a central role in decoupling and composition. In the following, we will elaborate details of each component.
\subsection{Model Architecture}\label{sec:kdgmodel}

\subsubsection{Knowledge-Grounded Response Generator}\label{sec:kdgmodel}

The generator $\mathcal{G}_G$ is a sequence-to-sequence model based on the Transformer architecture \citep{vaswani2017attention}, it consists of a 6-layers encoder and decoder with a hidden size of 768. Given a dialogue context $U = \{u_{1},\ldots, u_{i},\ldots,u_{l}\}$ with $u_{i}$ the $i$-th utterance, and a document $K = \{k_{1},\ldots,k_{i},\ldots,k_{h}\}$ with $k_{i}$ the $i$-th sentence. We concatenate $U$ and $K$  as a long sentence as the input of the encoder, then the decoder generates a response $\overline{Y}$ as output
\begin{align}
\overline{Y} = \{w_1, \ldots,w_i, \ldots ,w_m\} = \mathcal{G}_G(U,K) 
\end{align}

\subsubsection{Sequential Style Disentangler}\label{sec:template}

To identify and disentangle the style-related fragments from $\overline{Y}$, we employ a sequence labeling module named Sequential Style Disentangler $\mathcal{F}$ to model the probabilities $\{x_i\}_{i=1}^m$ of being style-related token at each position in $\overline{Y}$.
The formulations are as follows:
\begin{align}
P_{\mathcal{F}}(A|\overline{Y},U,K) &= \prod_{i=1}^{m} P(a_{i}|\overline{Y},U,K)\\
P(a_{i}|\overline{Y},U,K) &= x_{i} = {\rm sigmoid}({\boldsymbol{W}{e_{i}}}) \\
\{e_1,\ldots ,e_m\} &= {\rm BERT}(\overline{Y},U,K)
\end{align}
where $\boldsymbol{W}\in\mathbb{R}^{v \times 1}$ and $e_i\in\mathbb{R}^{v}$, $v$ is representation dimension, $a_i\in \{replace,retain\}$, $A = \{a_i\}_{i=1}^m$. Then when generating $\widetilde{Y}$ if $x_{i} > \varepsilon$, $a_i$ will be operation ``replace" indicating $w_i$ is a style token and needs to be replaced with a tag token [*], and viceversa for $x_{i} < \varepsilon$, $a_i$ will be operation ``retain" indicating $w_i$ remains unchanged. Threshold $\varepsilon$ is equal to top $P_r\%$ percentile of $\{x_i\}_{i=1}^m$, where $P_r$ is a hyper parameter. Finally, we perform the the predicted sequence of operations on $\overline{Y}$ to obtain style-agnostic template $\widetilde{Y}$. As the style disentangler tags each word in a sentence, it captures the style fragments (e.g., words, phrases, sub-sequences, or even the whole sentence) rather than only style tokens. The learning detail is presented in Appendix \ref{appendix:method}.


\subsubsection{Style Rewriter}

With $\widetilde{Y}$  as input, the style rewriter $\mathcal{G}_R$  generates a new  $\hat{Y}$  word-by-word in the target style. $\mathcal{G}_R$ has the same architecture as $\mathcal{G}_G$. The generation process of $\mathcal{G}_R$ is formulated as:
\begin{equation}
P_{R}(\hat{Y}|\widetilde{Y})=\prod_{t=1}^{h} P_{R}(\hat{w}_{t}|\widetilde{Y})
\end{equation}
where $\hat{w}_{t}$ is the $t$-th token of $\hat{Y}$ whose length is $h$.

\subsection{Reinforcement Learning}\label{sec:policy-learning}

Neither style disentangler nor style rewriter has supervision for training. Moreover, We need to ensure the correctness of $\hat{Y}$ without any modifications of the original content in the knowledge section of $\overline{Y}$. To cope with the challenges, we exploit REINFORCE~\citep{policygradient} to train $\mathcal{F}$ and $\mathcal{G}_R$ jointly with a total reward determined by the semantic similarity with the ground-truth response and the consistency with the desired style. Specifically, we maximize the expected reward as: 

\begin{align}
       \mathcal{R_{RL}} &=  \mathbb{E}_{\hat{Y} \sim P_{R}(\hat{Y})}\mathbb{E}_{\widetilde{Y} \sim P_{\mathcal{F}}(A)}[R(\widetilde{Y},Y)]
    \label{eq:pairwise-rl}
\end{align}
where $P_{R}(\hat{Y})$ and $P_{\mathcal{F}}(A)$ stand for $P_{R}(\hat{Y}|\widetilde{Y})$ and $P_{\mathcal{F}}(A|\overline{Y},U,K)$ respectively, $R(\widetilde{Y},Y) = {\rm{Sim}}(\hat{Y}, Y) +  {\rm{Cls}}(\hat{Y})$, ${\rm{Sim}}(\cdot)$ is embedding cosine similarity which supervises the knowledge regularization, ${\rm{Cls}}(\cdot)$ is the style intensity predicted by a classifier. We subtract the mean value of rewards $R$ in a batch to reduce the variance of gradient estimation~\citep{clark-manning-2016-deep}. In order to avoid the destroying issue of RL, we fix the parameters of $\mathcal{G}_R$, then only optimize the $\mathcal{F}$.

\subsection{Weakly Supervised Learning}

Since the style disentangler and style rewriter need to be carefully synchronized, ideally we hope they can benefit each other in learning. However, in the early stage as the parameters of $\mathcal{F}$ and $\mathcal{G}_R$ are far from optimal. It is possible that, on the one hand, the templates that are not decoupled successfully hinder $\mathcal{G}_R$ learning from rewriting style fragments accurately. On the other hand, noise signals from rewards computed with the low-quality responses generated by $\mathcal{G}_R$ flow to the learning of $\mathcal{F}$, resulting in inferior $\mathcal{F}$. To alleviate error accumulation in joint training, we propose a novel weakly supervised stylistic template disentangle method to assist the learning of $\mathcal{F}$ and $\mathcal{G}_R$.

\subsubsection{Weakly Supervised Disentangler}
\label{sec:disentangler_initialization}
\begin{figure}[ht]
\centering
     \includegraphics[width=0.55\textwidth, scale=1, trim=330 270 300 280,clip]{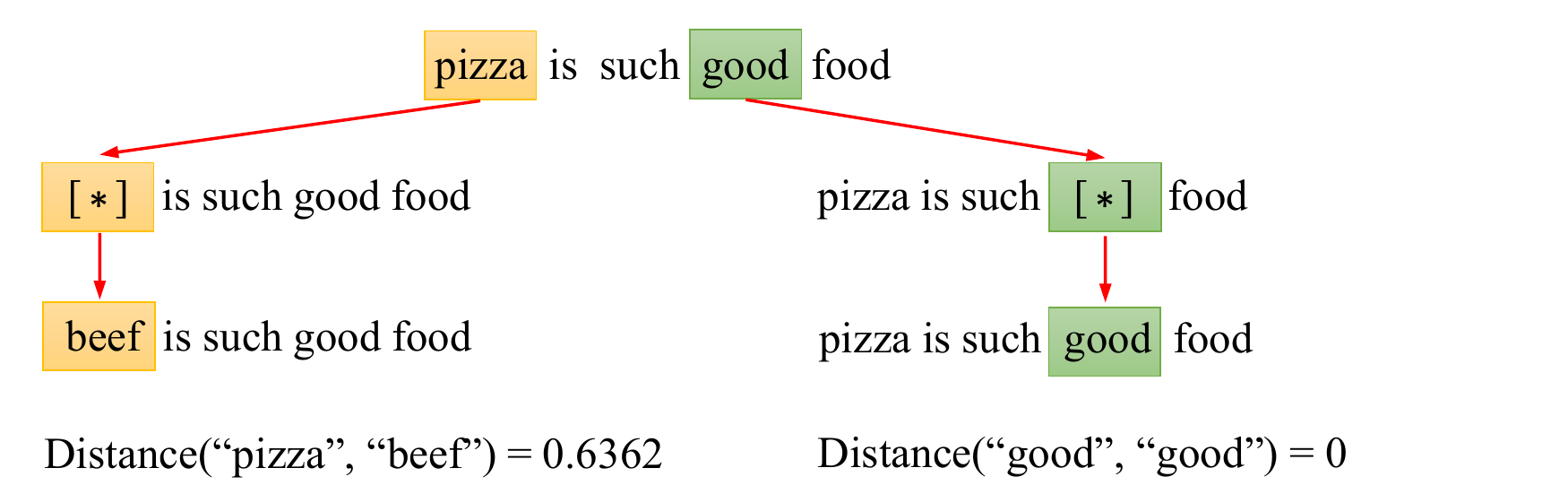}
     \caption{The positive sentiment word ``good'' is easier to be reconstructed than the knowledge word ``pizza'' in the sentence, where the wrong prediction ``beef'' would hurt the knowledge preservation and confuse the dialogue theme.}
     \label{fig:intuition_dae}
\end{figure}

Intuitively, style fragments dominate the distribution of style corpus $\mathcal{D}_{s}$ compared with content fragments, thus the style fragments are easier to be reconstructed than content fragments by the denoising autoencoder trained on $\mathcal{D}_{s}$. As shown in Figure \ref{fig:intuition_dae}, a denoising reconstruction model $\mathcal{G}_D$ reconstructs the style word ``good'' successfully but fail to do that for content word ``pizza'' in the same response from $\mathcal{D}_{c}$. Particularly, we randomly divide $\mathcal{D}_{s}$ into two halves with equal probability: $\mathcal{D}_{s}^1$ and $\mathcal{D}_{s}^2$, then $\mathcal{D}_{s}^1$ is used to train the denoising reconstruction model $\mathcal{G}_D$. The reconstruction objective $\mathcal{L}_\mathcal{S}$ is formulated as:
\begin{align}
\mathcal{{L}_\mathcal{S}}=\mathbb{E}_{T\sim {\mathcal{D}_{s}^1}}[-\log p(T|\widetilde{T})]\
\end{align}

\noindent{where $\widetilde{T}$ is the corrupted version of $T$ by randomly mask 15\% tokens.}

Then for each sentence $T = \{t_i\}_{i=1}^m$ (with $t_i$ the $i$-th token in $T$) in $\mathcal{D}_{s}^2$, we sequentially mask one token each time to construct its denoising versions $\{\widetilde{T}_i\}_{i=1}^m$, then $\{\widetilde{T}_i\}_{i=1}^m$ are inferenced by $\mathcal{G}_D$ to reconstruct $\{\hat{T}_i\}_{i=1}^m$. We acquire a distance sequence $\boldsymbol{d} = \{d_i\}_{i=1}^m = \{\operatorname{Dis}(t_i, \hat{t}_i)\}_{i=1}^m$ where $\operatorname{Dis}(\cdot, \cdot)$ denotes a distance function. 
Based on above intuition, lower $d_i$ means $t_i$ is more preferable as a style-related token, thus for $t_i$ and $t_j$, if $d_i < d_j$, we define the label $y = 1$, and viceversa for $d_i > d_j$, $y = -1$. We aggregate all $<t_i$, $t_j$, $y>$ triples to construct $\mathcal{D}_{s\_t}$ to optimize the style disentangler via the pairwise ranking loss:
\begin{align}
\mathcal{L}_\mathcal{P}(t_i, t_j, y)=\max{(0, -y * (t_i - t_j) + {\mu})}\
\label{eq:pairwise-loss}
\end{align}
where ${\mu}$ is a hyper parameter. 
The action space of token-level pairwise ranking is large, so for each sentence in $\mathcal{D}_{s\_t}$, we randomly sample $Z$ non-repetitive $<x_i, x_j, y>$ triples to optimize $\mathcal{L}_\mathcal{P}$, where Z is a hyper parameter. The  style tokens in various style corpus found by the style disentangler is presented in Appendix \ref{appendix:stawords}.

\subsubsection{Weakly Supervised Rewriter }\label{sec:rewriter_training}

The training data for the rewriter are also constructed by an unsupervised method: Optimized style disentangler $\mathcal{F}$ (Eq.\ref{eq:pairwise-loss}) infers the style corpus $\mathcal{D}_{s}= \{T_i\}_{i=1}^M$ and generates a disentangled template set ${\mathcal{\widetilde{D}}_{s}} = \{\widetilde{T}_i\}_{i=1}^M$. Then the rewriter takes paired  $< {{\widetilde{T}}}, {T}>$ as input and output respectively. Since $\widetilde{T}$ is style-agnostic, the rewriter would focus on transfering a factual sentence to a desired sentence with target style. The loss function for the rewriter $\mathcal{G}_R$ is:

\begin{equation}
    \label{eq:dialogpt-prob}
    \mathcal{L_R} = - \frac{1}{M} \sum_{l=1}^{M}(\prod^{|T_l|}_{i=1} p(t_{l,i} | t_{l,1},\cdots, t_{l,i-1}; \widetilde{T}_l))
\end{equation}
where $t_{l,i}$ is the $i$-th token in $l$-th sentence.  Specifically, the rewriter $\mathcal{G}_R$  has a same architecture as $\mathcal{G}_G$. 

\begin{algorithm}[htb]
\caption{Optimization Algorithm.}
\label{alg:optimize}
\begin{algorithmic}[1]

\STATE {\textbf{Input:} Datasets $\mathcal{D}_{c}$, $\mathcal{D}_{s}$; Models $\mathcal{G}_G$, $\mathcal{F}$,  $\mathcal{G}_R$.}

\STATE Optimize $\mathcal{G}_G$ using $\mathcal{D}_{c}$.\\
\STATE Construct $\mathcal{D}_{s\_t}$. \\
\STATE Optimize $\mathcal{F}$ using $\mathcal{D}_{s\_t}$  (Eq.\ref{eq:pairwise-loss}) . \\
\STATE Construct ${\mathcal{\widetilde{D}}_{s}}$ using $\mathcal{F}$. \\
\STATE Optimize $\mathcal{G}_R$ using $\mathcal{D}_{s}$ and ${\mathcal{\widetilde{D}}_{s}}$  (Eq.\ref{eq:dialogpt-prob}) . \\
\STATE Further Optimize $\mathcal{F}$  using $\mathcal{D}_{conv}$ (Eq.\ref{eq:pairwise-rl}) . \\

\RETURN $\mathcal{G}_G$, $\mathcal{F}$, $\mathcal{G}_R$.
\end{algorithmic}
\end{algorithm}

\section{Experiments}
We conduct experiments on Wizard of Wikipedia (Wizard) and Topical Chat with positive and negative sentiments, and polite style.


\subsection{Datasets}
\noindent\textbf{KDG Corpus} 
Wizard consists of 1365 topics, and each conversation happens between a wizard who has access to Wikipedia paragraphs and an apprentice who talks to the wizard. Topical Chat utilizes wiki articles, Washington Post, and Reddit fun facts as the knowledge source. The participants play symmetric and asymmetric roles according to the knowledge. Wizard and Topical Chat are split as training, valid and test set respectively. We compare our method with baselines on Wizard Test Seen and Topical Chat Test Freq. More details are described in Appendix \ref{appendix:data}.

\noindent\textbf{Style Corpus} We use Amazon dataset published in \citet{li2018amazom} and Politeness published in \citet{Aman2020polite} for style transfer. Amazon consists of product reviews from Amazon for flipping sentiment, and it contains 27800 positive sentences and 27700 negative sentences. For Politeness, We use the P9-bucket as the polite dataset, which consists of 27000 polite sentences.

\begin{table*}\scriptsize  
\renewcommand\arraystretch{1.2}
\renewcommand\tabcolsep{1.6pt}
\centering
\begin{tabular}{c|l |c| c c c c | c c c c| c|c | c c  c c |c  c c c|c }
\hline

\multirow{3}{*}{Style } &
\multirow{3}{*}{Models} &
\multicolumn{10}{c|}{Wizard of Wikipedia} &
\multicolumn{10}{c}{Topical Chat}  \\\cline{3-22}

& & {Style} &
\multicolumn{4}{c|}{Relevance} &
\multicolumn{4}{c|}{Diversity} &
{Average } &
{Style } &
\multicolumn{4}{c|}{Relevance} &
\multicolumn{4}{c|}{Diversity} &{Average } \\\cline{4-11} \cline{14-21}
  & & Intensity & F1&{B-1} & {B-2} &{R}  & {D-1} & {D-2} & {iD-1} & {iD-2} & {Length} &
  Intensity & F1 &{B-1} & {B-2}  &{R}  & {D-1} & {D-2} & {iD-1} & {iD-2} & {Length}\\

\hline

\multirowcell{4}{Positive} 
 &StyleFusion    &0.275 &11.8 &12.3 &6.5 &10.1 &3.3 &8.7 & 54.1 & 67.3 &11.2 &0.263 &12.6 &12.9 &6.8 &11.2 &0.8 &2.6 & 42.1 & 60.7 & 9.7\\
 &StylisticDLV   &0.336 &10.6 &11.5 &6.1 &9.3 &3.9 &9.2 & 56.7 & 69.5 &10.6 &0.381 &12.2 &12.5 &6.7 &10.6 &1.3 &3.3 & 44.7 & 63.2 & 10.3\\
 &StylizedDU    &0.342 &15.7 &17.4 &9.6 &18.5 & {14.1} & {34.8}& 50.3 & 65.5 &14.5  &0.417 &16.2 &15.8 &10.1 &15.4 &3.6 &10.5 & 46.2 & 65.8&12.8 \\
 &StyleDGPT &\textbf{0.354} &21.7 &24.3 &17.1 &24.8 &12.5 &33.2 & \textbf{61.4} & 73.2 & 10.8  &0.392 &20.4 &18.6 &14.6 &18.7 &2.8 &7.8 & 58.6 & 74.9&11.3  \\
& DTR    &0.338 & \textbf{31.3} & \textbf{32.6} & \textbf{20.7} & \textbf{32.6} &\textbf{12.9} &\textbf{35.5}& 59.6 & \textbf{76.9} &\textbf{20.3}  & \textbf{0.448} & \textbf{26.4} & \textbf{30.2} & \textbf{18.9} & \textbf{26.0} &\textbf{3.9} &\textbf{11.8} & \textbf{63.8} & \textbf{76.0}& \textbf{19.5} \\
\hline
\multirowcell{4}{Negative} 
&StyleFusion    &0.327 &12.5 &11.7 &7.4 &9.6 &3.1 &8.8 & 53.5 & 70.3& 10.4 &0.293 &10.8 &11.4 &6.5 &10.6 &1.0 &2.4 & 55.7 & 63.5&  10.9\\
 &StylisticDLV   &0.665 &11.8 &11.1 &6.9 &9.0 &3.4 &9.1 & 54.7 & 70.8 &11.3 &0.655 &10.4 &11.2 &6.1 &10.5 &1.2 &2.7 & 58.0 & 64.9 & 11.2\\
 &StylizedDU&0.640 &16.1 &16.7 &9.4 &15.9 &13.6 &31.3 & 56.1 & 69.6&13.8  &0.642 &15.7 &15.5 &11.3 &15.8 &3.2 &8.4 & 58.0 & 65.4& 12.5  \\
 &StyleDGPT &0.713 &22.5 &24.9 &17.5 &25.0 &11.8 &32.0 & 62.9 & 74.2&12.4  &0.686 &21.3 &22.1 &16.5 &19.2 &2.3 &6.6& 64.6 & 70.1 & 10.7    \\
& DTR   & \textbf{0.783} & \textbf{32.0} & \textbf{31.1} & \textbf{20.6} & \textbf{31.8} &\textbf{14.3} &\textbf{34.5} & \textbf{66.4} & \textbf{78.7}&\textbf{18.7}  & \textbf{0.715} & \textbf{27.9} & \textbf{31.2} & \textbf{19.7} & \textbf{26.5} &\textbf{4.5} &\textbf{12.8} & \textbf{67.2} & \textbf{75.3}& \textbf{21.2}  \\
\hline
\multirowcell{4}{Polite} 
&StyleFusion &0.211 &11.3 &11.6 &6.8 &10.7 &1.9 &5.5 & 45.0 & 53.4& 12.6 &0.243 &12.5 &12.8 &7.3 &12.2 &0.8 &2.3 & 40.4 & 57.1& 10.4  \\
 &StylisticDLV  &0.264 &10.7 &10.8 &6.2 &10.1 &2.1 &6.0 & 47.3 & 55.9 &12.1 &0.375 &13.0 &13.4 &7.5 &12.6 &0.9 &2.8 & 43.6 & 59.3 & 9.8\\
 &StylizedDU        &0.270 &14.9 &16.2 &10.2 &17.4 &11.5 &35.1 & 43.3 & 63.2& 14.7 &0.382 &16.4 &15.3 &10.9 &14.7 &3.8 &12.4 & 42.8 & 60.9& 13.9  \\
 &StyleDGPT       &0.262 &24.8 &22.2 &15.7 &23.8 &12.2 &33.1 & 51.9 & 65.7& 13.3 &0.316 &20.8 &19.4 &15.3 &20.8 &3.0 &9.2 & 45.7 & 58.3& 12.8  \\
& DTR   & \textbf{0.287} & \textbf{30.6} & \textbf{29.3} & \textbf{20.5} & \textbf{31.6} &\textbf{12.8} &\textbf{37.4}& \textbf{55.4} & \textbf{68.1} & \textbf{20.3} & \textbf{0.403} & \textbf{27.6} & \textbf{30.5} & \textbf{19.8} & \textbf{29.1} &\textbf{4.2} &\textbf{14.6} & \textbf{47.2} & \textbf{62.5}& \textbf{20.5}  \\
\hline

\end{tabular}
\caption{\label{font-table}  Automatic evaluation results. Numbers in bold
mean that the improvement to the best baseline is statistically significant (t-test with $p$-value $<$ 0.01).
}
\label{res:auto}
\end{table*}

\begin{table*}[hbt]\scriptsize
\setlength{\tabcolsep}{0.3mm}
\renewcommand\arraystretch{1.2}
\centering
\begin{tabular}{l | c | c c c | c c c | c c c  | c c c  | c | l | c | c c c | c}
\hline
\multicolumn{15}{c|}{Manual evaluation results}& \multicolumn{6}{c}{Attractiveness evaluation results}\\
\hline
\multirow{3}{*}{Style} &
\multirow{3}{*}{Models} &
\multicolumn{3}{c|}{Style} &
\multicolumn{3}{c|}{Knowledge} &
\multicolumn{3}{c|}{Context} & 
\multicolumn{3}{c|}{}& 
{}&
\multirow{3}{*}{Style} & 
\multirow{3}{*}{Models} &
\multicolumn{3}{c|}{} &
{}\\
& & \multicolumn{3}{c|}{Consistency} &
\multicolumn{3}{c|}{Preservation} &
\multicolumn{3}{c|}{Coherence} &
\multicolumn{3}{c|}{Fluency} &
{Kappa}& & & \multicolumn{3}{c|}{Attractiveness } &Kappa\\
  & &  W(\%) & L(\%) & T(\%) & W(\%) & L(\%) & T(\%) & W(\%) & L(\%) & T(\%) & W(\%) & L(\%) & T(\%) & & & & W(\%) & L(\%) & T(\%) & \\
\hline
\multicolumn{15}{c|}{Wizard of Wikipedia} & \multicolumn{6}{c}{Wizard of Wikipedia}\\
\hline
{Positive} 
 & DTR vs. StyleDGPT   &56.8 &22.2 &21.0     &58.0 &22.8 & 19.2  &52.4 &19.5 & 28.1  &54.9 &22.3 &22.8   &0.67 &{Positive} & DTR vs. DTR-s   &60.4 &18.1 &21.5 & 0.68 \\
\hline
{Negative} 
 & DTR vs. StyleDGPT    &54.8 &18.4 &26.8   &58.2 &17.9 & 23.9  &55.0 & 19.6& 25.4  &51.0 &28.6 & 20.4  &0.65 &{Negative} & DTR vs. DTR-s   &13.7 &58.3 & 28.0 & 0.65\\\hline
{Polite} 
 & DTR vs. StyleDGPT    &58.0 &21.6 & 20.4   &60.2 &21.1 & 18.7  &56.7 &20.7 & 22.6  &50.5 &29.2 & 20.3  &0.68&{Polite} & DTR vs. DTR-s    &56.2 &12.3 & 31.5 & 0.64 \\
\hline
\multicolumn{15}{c|}{Topical Chat} & \multicolumn{6}{c}{Topical Chat}\\
\hline
{Positive} & DTR vs. StyleDGPT    &48.2 &23.3 & 28.5  &57.3 &20.5 & 22.2  &48.8 &24.0 & 27.2  &53.6 &23.9 & 22.5  &0.65& {Positive} & DTR vs. DTR-s  &54.6 &23.1 & 22.3 & 0.65\\
\hline
{Negative} 
 & DTR vs. StyleDGPT    &56.7 &21.6 & 21.7   &51.6 &27.4 & 21.0  &54.0 &22.6 & 23.4  &52.6 &22.9 & 24.5  &0.64 &{Negative} & DTR vs. DTR-s    &26.4 &54.5& 19.1& 0.65\\\hline
{Polite} 
 & DTR vs. StyleDGPT    &49.8 &19.3 & 30.9  &46.5 &28.1 & 25.4  &45.6 &27.3 & 27.1  &53.5 &21.1 & 25.4  &0.65 &{Polite} & DTR vs. DTR-s  &49.6 &21.7 & 28.7  & 0.67\\
\hline
\end{tabular}
\caption{\label{font-table}  Manual evaluation results.  W, L, and T refer to Win, Lose, and Tie. All of the Kappa scores are greater than 0.6, which indicates the good agreement among the annotators. Other models are shown in Appendix \ref{appendix:manual}.
}
\label{res:human}
\end{table*}

\subsection{Evaluation Metrics}\label{sunsec:metircs}
Following \citet{zheng2020stylized} and \citet{yang2020styledgpt}, we use automatic metrics to measure DTR on three aspects: \textbf{Style Intensity}, \textbf{Relevance}, and \textbf{Diversity}. For style intensity, we use the GPT-2 classifier prediction mentioned in section \ref{sunsec:details}.  Relevance is measured with \textbf{F1}, \textbf{BLEU} \citep{papineni-etal-2002-bleu} and  \textbf{Rouge} \citep{lin-2004-rouge}. We use \textbf{Distinct} \citep{li-etal-2016-diversity} to measure Diversity of different models. To measure the diversity between different styles, we propose \textbf{inner Distinct}: given a context and knowledge, we calculate distinct in three generated responses with three styles.

For human evaluation, we randomly sample 500 examples from test set, and recruit 3 well-educated annotators. To each annotator, two responses from different models are presented, which are randomly shuffled to hide their sources. The annotators then judge which response is better from four aspects: (1) \textbf{Style Consistency}: which response exhibits the desired style more (2) \textbf{Knowledge Preservation}: which response is more relevant to the knowledgeable document (3) \textbf{Context Coherence}:  which response is more coherent with the dialogue context (4) \textbf{Fluency}: which response is more fluent and free from any grammar errors.

\subsection{Implementation Details}\label{sunsec:details}
We use pre-trained MASS \citep{song2019mass} to initialize $\mathcal{G}_G$ and $\mathcal{G}_R$. We adopt Adam optimizer as an initial learning rate of 5 $\times 10^{-4}$, and the batch size is 4096 tokens for a NVIDIA 1080 Ti GPU. Since all the baselines don't have a knowledge selection module, we chose the ground-truth knowledge as input for Wizard and the top-1 knowledge sentence according to the BLEU-1 with the corresponding response as input for Topical Chat. We use beam search(size=5) to decode the response.  We initialize $\mathcal{F}$ with pre-trained BERT, the replace rate $P_r$ is 25, $Z$ in section \ref{sec:policy-learning} is 10. We use Glove \citep{pennington2014glove} 100d embedding and cosine similarity as $\operatorname{Dis}(\cdot, \cdot)$ to calculate distance $d$.  ${\mu}$ in Eq.\ref{eq:pairwise-loss} is 0.2. To get the style intensity reward, we follow \citet{yang2020styledgpt} and train binary GPT-2 \citep{radford2019language} classifiers. Early stopping on validation is adopted as a regularization strategy. All the above hyperparameters are determined by grid search.

\begin{table*}[hbt]\rmfamily\scriptsize
\renewcommand\arraystretch{1.2}
\renewcommand\tabcolsep{2.5pt}
\centering
\begin{tabular}{l |c| c c c c | c c c c |c | c c  c c |c  c c c }
\hline

\multirow{3}{*}{Models} &
\multicolumn{9}{c|}{Wizard of Wikipedia} &
\multicolumn{9}{c}{Topical Chat}  \\\cline{2-19}

& {Style} &
\multicolumn{4}{c|}{Relevance} &
\multicolumn{4}{c|}{Diversity} &
{Style } &
\multicolumn{4}{c|}{Relevance} &
\multicolumn{4}{c}{Diversity}\\\cline{3-10} \cline{12-19}
  & Intensity & F1&{B-1} & {B-2} &{R}  & {D-1} & {D-2}& {iD-1} & {iD-2}  &
  Intensity & F1 &{B-1} & {B-2}  &{R}  & {D-1} & {D-2}& {iD-1} & {iD-2} \\
   
\hline
DTR    &0.338 & 31.3 & 32.6 & 20.7 & 32.6 &12.9 &35.5& 59.6 & 76.9  & 0.448 & 26.4 & 30.2 & 18.9 & 26.0 &3.9 &11.8 & 63.8 & 76.0  \\ \cline{1-19}
w/o  $\mathcal{F}$ WSL   &0.186 &23.4 &21.2 &13.5 &21.6 &15.1 &38.4 &67.3 & 80.4 &0.287 &15.5 &17.9 &11.6 &16.3 &4.8 &14.9  &69.1 & 81.3 \\
w/o $\mathcal{F}$ (TFIDF)  &0.244 &28.7 &28.5 &19.0 &29.6 &13.5 &36.6 &61.5 & 78.6 &0.369 &22.7 &26.2 &16.0 &21.5 &4.3 &12.3  &65.5 & 76.4 \\ 
 w/o $\mathcal{F}$ (Classification)   &0.256 &31.7 &30.2 &20.9 &31.5 &13.1 &35.3 &58.6 & 76.2 &0.375 &23.3 &26.7 &16.5 &22.6 &3.8 &12.1 &64.7 & 75.0 \\ \cline{1-19}
 w/o Rewards&0.307&31.4 &28.9 &20.2 &29.6 &12.7 &35.1 &58.1 & 76.4 &0.424 &25.1 &29.0 &18.3 &24.3 &4.2 &11.9  & 63.7& 74.2\\ 
 w/o Cls   &0.297 &32.0 &31.5 &20.8 & 30.2&12.0 &34.7 &56.3 & 74.3 &0.396 &24.4 &30.7 &19.1 &26.2 &3.9 &11.4  &63.1 & 74.5\\ 
w/o Sim  &0.340 &30.4 &28.6 &20.2 & 29.4&12.3 &36.5 &61.0 & 78.3 &0.452 &26.8 &28.8 &17.9 &24.3 &4.3 &12.1  & 64.8& 75.6\\

\hline
\end{tabular}
\caption{\label{font-table}  Ablation evaluation results of positive sentiment. Other styles are shown in Appendix \ref{appendix:ablation}.
}
\label{abl:ablation}
\end{table*}

\subsection{Baselines}
\begin{figure}[htb]
\centering
     \includegraphics[width=0.5\textwidth, scale=0.85, trim=15 6 0 37,clip]{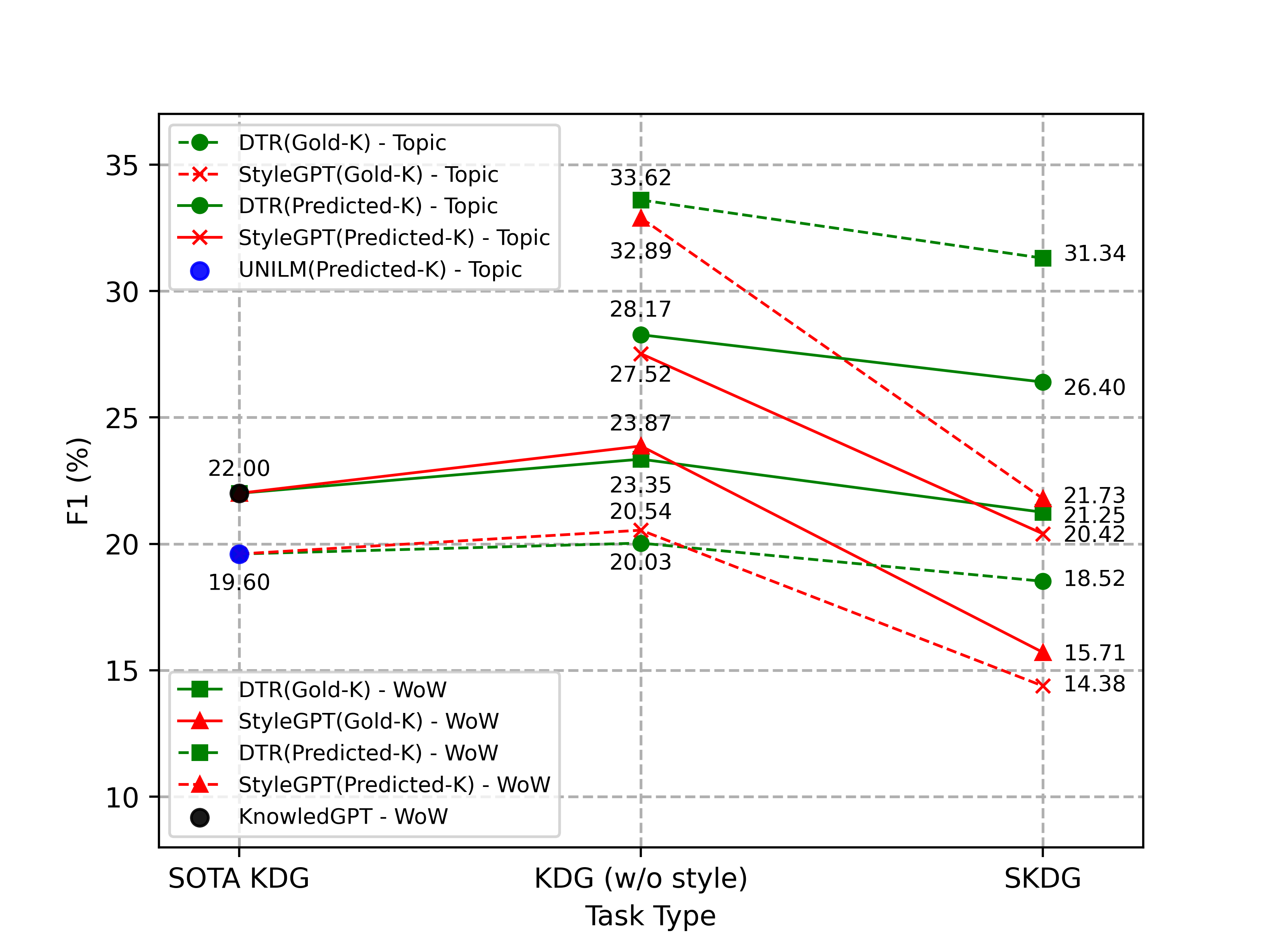}
     \caption{F1 of DTR and StyleDGPT (positive), and SOTA KDG models in different evaluation settings. }
     \label{fig:drop}
\end{figure}
The following models are selected as baselines: 

\begin{figure}[ht]
\centering
     \includegraphics[width=0.5\textwidth, scale=0.85, trim=15 6 0 37,clip]{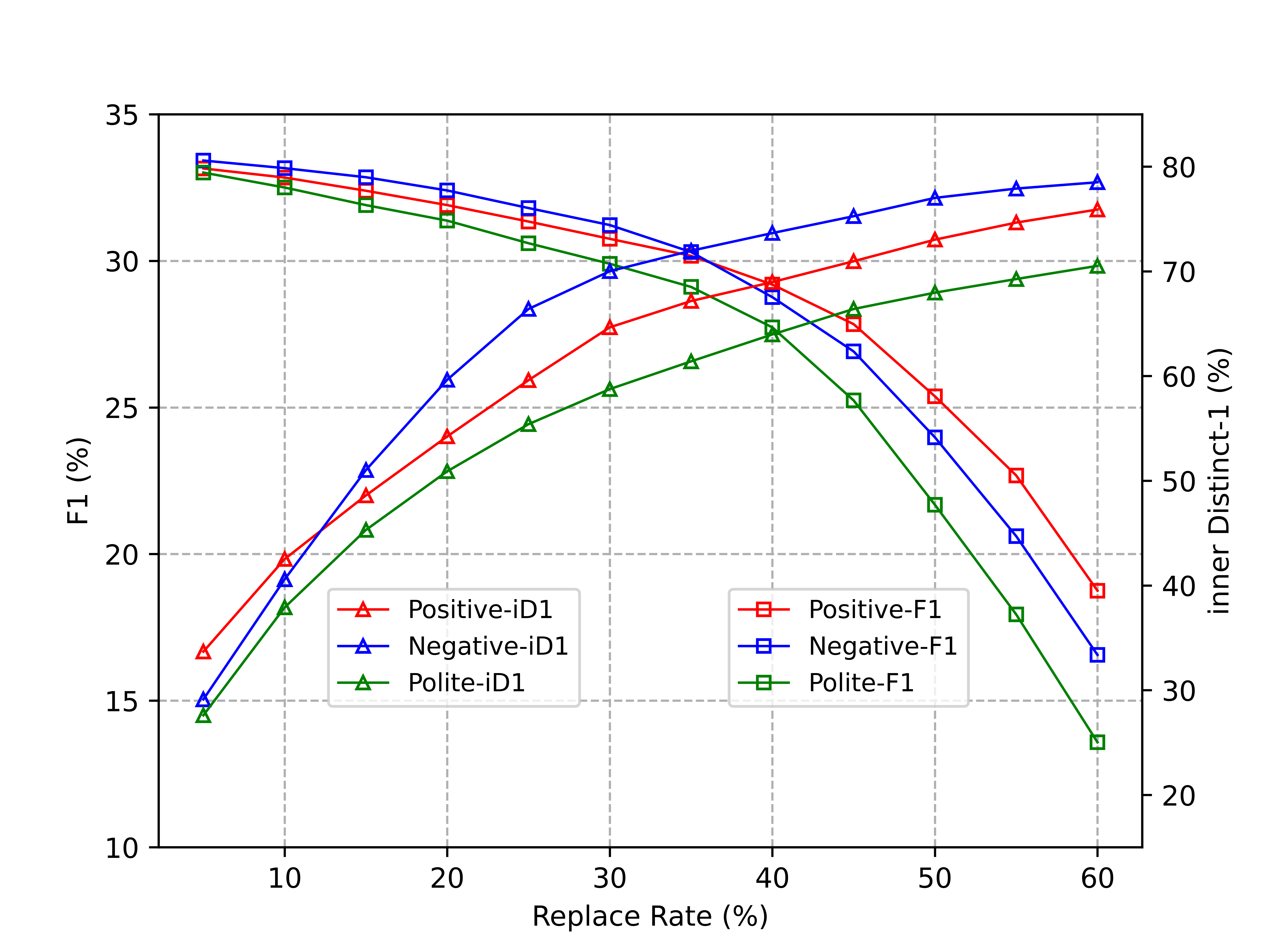}
     \caption{F1 and inner Distinct with different replace rate on three different styles in Wizard Test set.}
     \label{fig:mask}
\end{figure}
\begin{itemize}
\item\textbf{StyleFusion} \citep{gao2019stylefusion} bridges conversation modeling and nonparallel style transfer by sharing a latent space. We use the code {\url{https://github.com/golsun/StyleFusion}}.
\item\textbf{StylisticDLV} \citep{zhu-etal-2021-neural} disentangles the content and style in latent space by diluting information in style representations. We use the code {\url{https://github.com/golsun/StyleFusion}}.
\item\textbf{StylizedDU} \citep{zheng2020stylized} leverages back-translation technique to generate pseudo stylized context-response pairs. We use the code {\url{https://github.com/silverriver/Stylized_Dialog}}.
\item\textbf{StyleDGPT}  \citep{yang2020styledgpt} exploits the pre-trained language models on the stylized response generation task. We use the code {\url{https://github.com/TobeyYang/StyleDGPT}}. 
\end{itemize}





All the baselines are jointly learned with datasets  $\mathcal{D}_{c}$ and  $\mathcal{D}_{s}$, and take the concatenation of knowledge and context as input.

\subsection{Evaluation Results}
As shown in Table \ref{res:auto}, our DTR model achieves competitive performance in style transfer and significantly outperforms the baselines in all the relevance metrics. This indicates that DTR can produce high-quality responses which are coherent to the context, related to the knowledge, and consistent with the target style simultaneously. We also observe that all SDG methods frequently lost the knowledge part (Appendix \ref{appendix:case}). DTR significantly outperforms StyleDGPT on relevance, indicating that leveraging the style intensity score to optimize the decoupling of the template is superior to directly optimizing response generation (degrading language modeling). We observe the core component back-translation in StylizedDU fails to infer pseudo-knowledge from a response (generally, the knowledge carries much more information than the response). Table \ref{res:human} reports the results of human evaluation, DTR significantly outperforms StyleDGPT on all aspects. DTR is also superior to all the baselines as the \textbf{Case Study} section  in Appendix \ref{appendix:case}.

\subsection{Ablation Study}


Firstly, to verify the contributions of the proposed disentangler and weakly supervised learning method, we consider the following variants:
(1) \textbf{w/o $\mathcal{F}$ WSL}: training DTR without the \textbf{W}eakly \textbf{S}upervised \textbf{L}earning of $\mathcal{F}$ in section \ref{sec:disentangler_initialization}. (2) \textbf{ w/o $\mathcal{F}$ (Classification)}: replace the pairwise ranking loss in $\mathcal{F}$ with a binary classification loss. We define those tokens with $d=0$ (in section \ref{sec:disentangler_initialization}) as style words (label=1), otherwise non-style words (label=0). (3) \textbf{w/o $\mathcal{F}$ (TFIDF)}: replace $\mathcal{F}$ with a TFIDF-based rule (replace the fragments as [*] in a sentence with the lowest $P_r$\% TFIDF scores except stop words). Table \ref{abl:ablation} shows the results of the three variants. We can conclude that (1) the weakly supervised learning of $\mathcal{F}$ is crucial to training DTR, since the variant with a simple TFIDF significantly outperforms the one without any initialization; and (2) the ranking loss in $\mathcal{F}$ plays a key role in the success of style transfer, there is a dramatic drop on the style intensity  of \textbf{ w/o $\mathcal{F}$ (Classification)}. According to our observation, it is overfitting on the style corpus, leading to a low success rate.


Secondly, to investigate the RL rewards in Eq.\eqref{eq:pairwise-rl}, we consider the following variants: (1) \textbf{w/o Rewards} : remove the similarity and style intensity reward. (2) $\textbf{w/o Sim}$: remove the similarity reward. (3) $\textbf{w/o Cls}$: remove the style intensity reward. As shown in Table \ref{abl:ablation}, removal any of the two rewards will cause performance drop, indicating that style intensity and similarity reward can enhance DTR. We also add $\textbf{Sim}$ to StylizedDU, the improvement is +2.1 on F1, thus it's hard for $\textbf{Sim}$ to bridge the huge gap. Negative and Polite are similar, these results are presented in Appendix \ref{appendix:ablation}.

\subsection{Discussions}

\noindent\textbf{Impact of stylized knowledge-grounded generation.} We annotate the ``Attractiveness'' (the annotators are given two different responses with the same context and knowledge from two different models, and they should determine which response is more attractive and engaging in a holistic way) of DTR and DTR-s (without style transfer) following the same process in \ref{sunsec:metircs}. Table \ref{res:human} reports the evaluation results. We can see that introducing a positive sentiment or a polite style would enhance the engagement of the KDG model while establishing a negative sentiment harm the attractiveness.


\noindent\textbf{Impact of style transfer on the conversational ability of SDG models.} We are curious about to what extent the conversational ability of SDG models will be damaged after style transfer. We examine DTR and StyleDGPT in two settings: (1) Gold-K: the given knowledge is the ground-truth (2) Predicted-K: the given knowledge is selected from a knowledge selection model \citep{zhao2020knowledgpt}. As shown in Figure \ref{fig:drop}, after style transfer on Wizard, the F1 of DTR drops 2.28 and 2.1 in Gold-K and Predicted-K, while the F1 of StyleDGPT drops 11.16 and 8.16 respectively. On Topical Chat,the F1 of DTR drops 1.77 and 1.51 in Gold-K and Predicted-K, while the F1 of StyleDGPT drops 7.1 and 6.16 respectively. Compared with StyleDGPT, DTR dramatically reduces the damage to the conversational ability while achieving a high success rate of style transfer. Thanks to the superior style transferring mechanism, our DTR achieves comparable performance with the state-of-the-art KDG models $\{$KnowledGPT\cite{zhao2020knowledgpt} on Wizard, UNILM\cite{li2020zero} on Topical Chat$\}$ in the standard KDG evaluation setting even after style transfer. The results of Negative and Polite are similar and presented in Appendix \ref{appendix:f1}.


\noindent\textbf{Impact of the replace rate $P_r$.} As shown in Figure \ref{fig:mask}, $P_r$ = 25 achieves the best balance between relevance and diversity. A smaller $P_r$ would remain a large number of original style fragments in the template, leading to tiny differences between different styles. On the contrary, a larger $P_r$ would delete those content fragments, which are harder to restore by rewriter, but the responses from different styles will be more diverse. Topical Chat follows the same regularity as shown in Appendix \ref{appendix:replace}.  

\section{Conclustion}

We explore stylized knowledge-grounded dialogue generation by proposing bridging the knowledge-grounded response generation with the stylized rewriting via sharing a  disentangled template.  Evaluation results on benchmarks of the task indicate that our model can achieve state-of-the-art performance and exhibits a superior generation ability over different knowledge domains and styles.

\section*{Acknowledgement}
We thank anonymous reviewers for their insightful suggestions to improve this paper.
\bibliography{anthology,custom}
\bibliographystyle{acl_natbib}

\clearpage
\appendix

\section{Method Detail} 
\subsection{Disentangler BERT Learning Detail} \label{appendix:method}

We initialize $\mathcal{F}$ with two pre-trained BERT. In the unsupervised initialization stage we only train the BERT$^{\alpha}$, then in the reinforcement learning stage, we fix the parameters of BERT$^{\alpha}$ and loosen BERT$^{\beta}$.

\begin{align}
P_{\mathcal{F}}(A|\overline{Y},U,K) &= \prod_{i=1}^{m} P(a_{i}|\overline{Y},U,K)\\
P(a_{i}|\overline{Y},U,K) &= x_{i} = x^{\alpha}_{i} + x^{\beta}_{i}\\ \label{eq:eqp}
x^{\alpha}_{i} &= {\rm sigmoid}({\boldsymbol{W^{\alpha}}{e^{\alpha}_{i}}}) \\
x^{\beta}_{i} &= {\rm sigmoid}({\boldsymbol{W^{\beta}}{e^{\beta}_{i}}}) \\
\{e^{\alpha}_1,\ldots ,e^{\alpha}_m\} &= {\rm BERT^{\alpha}}(\overline{Y})\\
\{e^{\beta}_1,\ldots ,e^{\beta}_m\} &= {\rm BERT^{\beta}}(\overline{Y},U,K)
\end{align}

\section{Experiments}
\subsection{Datasets} \label{appendix:data}
Table \ref{data:datasets} reports the statistics of the Wizard of Wikipedia dataset and the Topical Chat dataset.

\begin{table*}[hbt]\scriptsize  

\renewcommand\arraystretch{1.2}
\centering
\begin{tabular}{c | c  c c c | c c c c c}
\hline
\multirow{2}{*}{Dataset } &
\multicolumn{4}{c|}{Wizard of Wikipedia} &
\multicolumn{5}{c}{Topical Chat} \\\cline{2-10}
   & Train & Valid & Test Seen & Test Unseen & Train & Valid Freq. & Valid Rare & Test Freq.& Test Rare\\
\hline
 Utterances  & 166787 & 17715 & 8715 & 8782 & 188378	& 11681 & 	11692	& 11760	& 11770 \\
\hline
 Conversations     & 18430 & 1948 & 965 & 968 & 8628& 	539	& 539 &	539	& 539 \\
\hline
 Average Turns & 9.0 & 9.1 & 9.0 & 9.1 & 21.8& 	21.6& 	21.7& 	21.8& 	21.8 \\
\hline
\end{tabular}
\caption{\label{font-table} Statistics for Wizard of Wikipedia, Topical Chat datasets.
}
\label{data:datasets}
\end{table*}

\subsection{Case Study} \label{appendix:case}

Table \ref{case:wizard}  and Table \ref{case:topical} presents some examples from Wizard of Wikipedia and Topical Chat respectively. In each case, we show the dialogue context, the knowledge (ground-truth), the human response, and responses from different models with each style.  We can see that responses from DTR  and StyleDGPT are well grounded by the provided knowledge and have obvious style , while responses from StyleFusion and StylizedDU in general lack of both informative content and desired style.  Compared with StyleDGPT, DTR is better at leveraging target style in the test phase and replies with more informative and more contextually coherent responses, which demonstrates the potential of the model in practice. For DTR, the knowledge-grounded response generator $\mathcal{G}_G$ firstly generates a factual response with mixed style-related tokens (such as ``yeah'', ``like'', ``not'', ``whether'', etc.) and content, then the template generator $\mathcal{F}$ replace them with a tag token [*] to produce a disentangled template, finally the rewriter  $\mathcal{G}_R$ modifies the tag [*] to generate some new sentences  in different target styles.

\begin{table*}[hbt]\small  
\renewcommand\arraystretch{1.12}
\centering
\begin{tabular}{p{48pt}p{52pt}| p{330pt} }
\hline
\multicolumn{2}{l|}{Knowledge} & a \textcolor{purple}{ grilled cheese sandwich} is made by grilling the sandwich with \textcolor{purple}{butter} or toasting it .  \\
\hline
\multirow{6}{*}{Context}& & A: hot dog . i love a good hotdog !  \\
& & B:  archery is a sport/skill of using a bow to propel arrows and a great sport it is . \\
& & A:   do you know where archery originated from ? 
 \\
& & B:   it's a delicious sausage sandwich . add a little mustard to it and a coke and that's a fine meal  \\
& & A: absolutely ! need to get me some homemade mustard plants .     \\
& & B:  lol ! what other quick meals do you like ? for example grilled cheese with chips ?    \\
\hline
Human &  &  i love butter on my grilled cheese !\\\hline
&$\mathcal{G}_G$ & \textcolor{red}{yeah}, i \textcolor{red}{like} the grilled cheese sandwich with butter.  \\
& $\mathcal{F}$ & \textcolor{red}{[*]}, i \textcolor{red}{[*]} the grilled cheese sandwich with butter.  \\
\multirow{4}{*}{Positive}
& DTR  & \textcolor{blue}{certainly}, i \textcolor{blue}{enjoy} the \textcolor{blue}{delicious} \textcolor{purple}{butter} on \textcolor{purple}{grilled cheese sandwich }.\\ \cline{2-3}
& StyleFusion  & yes, i think so too. \\
& StylizedDU  &   yes, i heard about the \textcolor{purple}{cheese}. \\
&  StyleDGPT &   i like \textcolor{purple}{toasting the sandwich}.  \\
\hline
\multirow{4}{*}{Negative}& DTR & \textcolor{blue}{I don't think so }, i \textcolor{blue}{hate} the \textcolor{purple}{grilled cheese sandwich} with \textcolor{blue}{greasy} \textcolor{purple}{butter}. \\ \cline{2-3}
& StyleFusion   &  i hate the other quick meals.\\
&  StylizedDU &  i did not know that. what is it about? \\
&  StyleDGPT & i don't know. I think it would be a bad idea. \\
\hline
\multirow{4}{*}{Polite}& DTR&  \textcolor{blue}{i am so sorry}, i ate a little \textcolor{purple}{grilled cheese sandwich with butter}. \\ \cline{2-3}
& StyleFusion & you know i am a big fan of \textcolor{purple}{cheese sandwich}.\\
& StylizedDU &  i don't know. I think it would be a bad idea.\\
& StyleDGPT  &  thanks for your \textcolor{purple}{grilled cheese sanwich}. \\
\hline
\end{tabular}
\caption{\label{font-table} Case study of Wizard of Wikipedia. Style-related words discovered by the style decoupler are marked in the \textcolor{red}{red} color, the generated style-related words of DTR are marked in the \textcolor{blue}{blue} color, and the  knowledge related words of all baselines and our model are marked in the \textcolor{purple}{purple} color.
}
\label{case:wizard}
\end{table*}

\begin{table*}[hbt]\small  
\renewcommand\arraystretch{1.2}
\centering
\begin{tabular}{p{50pt}p{52pt}| p{320pt} }
\hline
\multicolumn{2}{l|}{Knowledge} & \textcolor{purple}{Former Partiots RB BenJarvus Green-Ellis} has never fumbled the football in his NFL career. \\
\hline
& & A: cold bench. Then again, I wouldn't want to be some place that cold or watching football.\\
& & B:  I'd rather watch it inside where it's warm.  Have you heard about the Georgia Tech-Cumberland game of 1916? \\
\multirow{8}{*}{Context}& & A:  No, what happened in that game? \\
& & B:   Georgia Tech defeated Cumberland but here's the thing, they defeated them by a score of 222-0!  \\
& & A:  That is insane. How could that even happen?  \\
& & B:   I don't know but it did.  It's the highest scoring game in history. \\
& & A:   I'm sure. I don't even watch much and I couldn't imagine that score. I wonder if most people left or were they curious to see how high it would go? \\
& & B: I guess it depended on what team you were pulling for.  To me, it's surprising that the highest scoring game was in college football and not professional. \\
& & A:   Maybe it is because some are not as good in college so they may be playing against someone not on their level.\\ 
& & B:   Good point.  Professional does have a player that has never fumbled the ball. \\
\hline
Human &  & I've heard that. Wasn't it a Patriot player? \\\hline
&$\mathcal{G}_G$ &  i am \textcolor{red}{not} sure \textcolor{red}{whether} he was benjarvus green-ellis.\\
& $\mathcal{F}$   & i am \textcolor{red}{[*]} sure \textcolor{red}{[*]} he was benjarvus green-ellis.\\
\multirow{4}{*}{Positive}
& DTR  & i am \textcolor{blue}{pretty} sure, because \textcolor{blue}{i am the loyal fan of}  \textcolor{purple}{benjarvus green-ellis}.  \\ \cline{2-3}
& StyleFusion  & i think it's funny that \textcolor{purple}{green-ellis} has never fumbled the football.  \\
& StylizedDU  &  that's impressive. \\
&  StyleDGPT &  i agree, the player was \textcolor{purple}{former partiots rb benJarvus}. \\
\hline
\multirow{4}{*}{Negative}& DTR & i \textcolor{blue}{don't know whether } he was \textcolor{purple}{benjarvus green-ellis as a former partiots rb}. \\ \cline{2-3}
& StyleFusion   &   are you a football fan?\\
&  StylizedDU &  \textcolor{purple}{green-ellis} has never fumbled the football.\\
&  StyleDGPT &  no, i didn't know about nfl. \\
\hline
\multirow{4}{*}{Polite}& DTR&  i am sure \textcolor{blue}{and please note that} he was \textcolor{purple}{benjarvus green-ellis}. \\ \cline{2-3}
& StyleFusion & i also saw the nfl this year. \\
& StylizedDU & i hope i never fumbled the football.\\
& StyleDGPT  & could you please tell me who is the player? \\
\hline
\end{tabular}
\caption{\label{font-table} Case study of Topical Chat. Style-related words discovered by the style decoupler are marked in the \textcolor{red}{red} color, the generated style-related words of DTR are marked in the \textcolor{blue}{blue} color, and the  knowledge related words of all baselines and our model are marked in the \textcolor{purple}{purple} color.
}
\label{case:topical}
\end{table*}

\begin{table*}[hbt]\rmfamily\scriptsize
\renewcommand\arraystretch{1.1}
\renewcommand\tabcolsep{2.5pt}
\centering
\begin{tabular}{c|l |c| c c c c | c c c c |c | c c  c c |c  c c c }
\hline

\multirow{3}{*}{Style } &
\multirow{3}{*}{Models} &
\multicolumn{9}{c|}{Wizard of Wikipedia} &
\multicolumn{9}{c}{Topical Chat}  \\\cline{3-20}

& & {Style} &
\multicolumn{4}{c|}{Relevance} &
\multicolumn{4}{c|}{Diversity} &
{Style } &
\multicolumn{4}{c|}{Relevance} &
\multicolumn{4}{c}{Diversity}\\\cline{4-11} \cline{13-20}
  & & Intensity & F1&{B-1} & {B-2} &{R}  & {D-1} & {D-2}& {iD-1} & {iD-2}  &
  Intensity & F1 &{B-1} & {B-2}  &{R}  & {D-1} & {D-2}& {iD-1} & {iD-2} \\
   
\hline
\multirowcell{4}{Positive} 
& DTR    &0.338 & 31.3 & 32.6 & 20.7 & 32.6 &12.9 &35.5& 59.6 & 76.9  & 0.448 & 26.4 & 30.2 & 18.9 & 26.0 &3.9 &11.8 &63.8 & 76.0  \\ \cline{2-20}
& w/o $\mathcal{F}$ Initialization  &0.186 &23.4 &21.2 &13.5 &21.6 &15.1 &38.4 &67.3 & 80.4 &0.287 &15.5 &17.9 &11.6 &16.3 &4.8 &14.9  &69.1 & 81.3 \\
& w/o $\mathcal{F}$ (TFIDF)  &0.244 &28.7 &28.5 &19.0 &29.6 &13.5 &36.6 &61.5 & 78.6 &0.369 &22.7 &26.2 &16.0 &21.5 &4.3 &12.3  &65.5 & 76.4 \\ 
& w/o $\mathcal{F}$ (Classification)   &0.256 &31.7 &30.2 &20.9 &31.5 &13.1 &35.3 &58.6 & 76.2 &0.375 &23.3 &26.7 &16.5 &22.6 &3.8 &12.1 &64.7 & 75.0 \\ \cline{2-20}
& w/o Rewards&0.307&31.4 &28.9 &20.2 &29.6 &12.7 &35.1 &58.1 & 76.4 &0.424 &25.1 &29.0 &18.3 &24.3 &4.2 &11.9  & 63.7& 74.2\\ 
& w/o Cls   &0.297 &32.0 &31.5 &20.8 & 30.2&12.0 &34.7 &56.3 & 74.3 &0.396 &24.4 &30.7 &19.1 &26.2 &3.9 &11.4  &63.1 & 74.5\\ 
& w/o Sim  &0.340 &30.4 &28.6 &20.2 & 29.4&12.3 &36.5 &61.0 & 78.3 &0.452 &26.8 &28.8 &17.9 &24.3 &4.3 &12.1  & 64.8& 75.6\\
\hline
\multirowcell{4}{Negative} 
& DTR &  0.783 & 32.0 & 31.1 & 20.6 & 31.8 &14.3 &34.5 & 66.4 & 78.7  & 0.715 & 27.9 & 31.2 & 19.7 & 26.5 &4.5 &12.8 & 67.2 & 75.3 \\\cline{2-20}
& w/o $\mathcal{F}$ Initialization  &0.508 &21.7 &20.9 &13.8 &23.5 &17.2 &39.7 &68.8 & 81.7 &0.425 &16.8 &16.3 &10.1 &14.2 &5.4 &14.3  &69.6 & 77.0 \\
& w/o $\mathcal{F}$ (TFIDF)   &0.727 &30.1 &29.7 &18.9 & 28.7 &14.8 &34.7 &68.5 & 79.1 &0.647 &25.7 &29.3 &18.0 &25.4 &4.9 &12.4  &66.4 & 74.0\\ 
& w/o $\mathcal{F}$ (Classification)   &0.705 &31.0 &30.6  &20.6 &31.0  &15.0 &35.1 &67.9 & 79.6 &0.633 &26.2 &30.2 &18.4 &25.7 &5.1 &13.3  &68.3 &75.8 \\\cline{2-20}
& w/o Rewards     &0.768 &30.9 &30.1 &20.2 &30.6 &14.9 &35.4  &66.8 &79.9 &0.698 &27.1 &30.4 &18.1 &25.3 &5.2 &12.6 &67.0 & 75.6 \\
& w/o Cls   &0.759 &32.1 &31.2 &21.4 &32.1 &13.8 &34.6  &64.3 &78.3 &0.687 &28.0 &31.5 &20.0 &26.9 &4.1 &11.9  &66.1 & 73.2\\ 
& w/o Sim        &0.786 &30.4 &29.9 &19.7 &30.5 &15.2 &35.9 &68.9 & 80.8 & 0.720&26.1 &30.6 &19.3 &26.5 &5.5 &12.7  &68.5 & 77.3\\
\hline
\multirowcell{4}{Polite} 
& DTR  &  0.287 & 30.6 & 29.3 & 20.5 & 31.6 &12.8 &37.4& 55.4 & 68.1 & 0.403 & 27.6 & 30.5 & 19.8 & 29.1 &4.2 &14.6 & 47.2 & 62.5  \\ \cline{2-20}

& w/o $\mathcal{F}$ Initialization  &0.156 &22.9 &20.5 &11.7 &19.6 &14.9 &40.6 &59.8 & 72.3 &0.282 &16.1 &18.3 &12.7 &17.5 &5.3 &16.9  &55.8 & 70.1 \\

& w/o $\mathcal{F}$ (TFIDF)  &0.214 &27.0 &27.8 &18.8 &29.8 &13.0 &38.1 &56.3 & 69.3&0.341 &23.1 &28.0 &18.2 &26.9 &4.9 &15.5 &48.7 & 63.5 \\ 
& w/o $\mathcal{F}$ (Classification)  &0.258 &30.9 &30.2 &21.3 &32.4&12.2 &36.6 &52.1 & 67.0 &0.375 &25.9 &29.4 &18.6 &27.1 &4.0 &14.8 &47.6 &62.8  \\ \cline{2-20}
& w/o Rewards&0.266 &31.0 &27.8 &20.2 &31.7 &12.6 &37.6  & 55.9&68.5 &0.384 &26.1 &29.1&19.1 &27.1 &4.3 &15.1 & 47.3& 63.6 \\ 
& w/o Cls   &0.265 &32.9 &30.6 &21.3 &32.3 &11.9 &37.0  &53.6 &67.6 &0.379 &27.8 &31.1 &20.1 &29.3 &3.8 &14.3 &45.5 &  62.4\\ 
& w/o Sim    &0.292 &30.7 &27.5 &20.0 &31.2 &13.1 &37.2  &56.3 &69.7 & 0.406&26.9 &28.9 &18.7 &27.8 &4.6 &15.2 & 48.0& 65.1 \\
\hline
\end{tabular}
\caption{\label{font-table}  Ablation evaluation results.
}
\label{abl:ablation}
\end{table*}

\subsection{Ablation evaluation} \label{appendix:ablation}
As shown in table \ref{abl:ablation}, we list all ablation evaluation results of Positive, Negative and Polite on Wizard of Wikipedia and the Topical Chat.

\subsection{Manual evaluation} \label{appendix:manual}
As shown in Table \ref{font-table}, we list all manual evaluation results of Positive, Negative and Polite on Wizard of Wikipedia and the Topical Chat.
\begin{table*}[hbt]\scriptsize
\setlength{\tabcolsep}{0.9mm}
\renewcommand\arraystretch{1.1}
\centering
\begin{tabular}{c | c | c c c | c c c | c c c | c c c  | c}
\hline
\multirow{3}{*}{Style} &
\multirow{3}{*}{Models} &
\multicolumn{3}{c|}{Style} &
\multicolumn{3}{c|}{Knowledge} &
\multicolumn{3}{c|}{Context} & 
\multicolumn{3}{c|}{Fluency}& 
{Kappa}\\
& & \multicolumn{3}{c|}{Consistency} &
\multicolumn{3}{c|}{Preservation} &
\multicolumn{3}{c|}{Coherence} &
\multicolumn{3}{c|}{} &
{}\\
   & &  W(\%) & L(\%) & T(\%) & W(\%) & L(\%) & T(\%) & W(\%) & L(\%) & T(\%) & W(\%) & L(\%) & T(\%) \\
\hline
\multicolumn{15}{c}{Wizard of Wikipedia} \\
\hline
\multirowcell{3}{Positive} 
 & DTR vs. StyleFusion  &53.6 &17.1 &29.3    &66.3 &10.5 & 23.2  &54.1 &10.4 & 35.5  &53.5 &21.8 & 24.7  &0.72  \\
 & DTR vs. StylizedDU   &57.4 &24.9 &17.7     &59.1 &24.1 & 16.8  &46.0 &23.2 & 30.8  &50.8 &23.1 & 26.1  &0.69 \\
 & DTR vs. StyleDGPT    &56.8 &22.2 &21.0      &58.0 &22.8 & 19.2  &52.4 &19.5 & 28.1  &54.9 &22.3 &22.8   &0.67 \\

\hline
\multirowcell{3}{Negative} 
& DTR vs. StyleFusion   &59.7 &22.9 &17.4     &65.7 &15.9 & 18.4  &58.0 &16.7 & 25.3  &55.9 &18.2 &25.9   &0.68 \\
 & DTR vs. StylizedDU   &57.2 &24.0 &18.8     &57.9 &23.0 & 19.1  &50.1 & 20.9& 29.0  &46.5 &24.8 & 28.7  &0.66 \\
 & DTR vs. StyleDGPT    &54.8 &18.4 &26.8     &58.2 &17.9 & 23.9  &55.0 & 19.6& 25.4  &51.0 &28.6 & 20.4  &0.65 \\
\hline
\multirowcell{3}{Polite} 
& DTR vs. StyleFusion   &60.9 &15.9 &23.2     &64.3 &7.3  & 28.4  &55.3 &16.1 & 28.6  &47.1 &25.2 & 27.7  &0.70 \\
 & DTR vs. StylizedDU   &58.7 &22.1 &19.2     &58.6 &20.4 & 21.0  &47.8 &21.2 & 31.0  &45.6 &31.8 & 22.6  &0.66 \\
 & DTR vs. StyleDGPT     &58.0 &21.6 & 20.4    &60.2 &21.1 & 18.7  &56.7 &20.7 & 22.6  &50.5 &29.2 & 20.3  &0.68 \\
\hline
\multicolumn{15}{c}{Topical Chat} \\
\hline
\multirowcell{3}{Positive} 
& DTR vs. StyleFusion   &54.8 &16.5 & 28.7    &53.0 &13.6 & 33.4  &56.0 &17.2 & 26.8  &53.5 &17.4 & 29.1  &0.69 \\
 & DTR vs. StylizedDU   &45.9 &19.4 & 34.7    &49.1 &21.3 & 29.6  &52.7 &21.4 & 25.9  &46.7 &20.2 & 33.1  &0.63 \\
 & DTR vs. StyleDGPT    &48.2 &23.3 & 28.5    &57.3 &20.5 & 22.2  &48.8 &24.0 & 27.2  &53.6 &23.9 & 22.5  &0.65 \\

\hline
\multirowcell{3}{Negative} 
& DTR vs. StyleFusion   &56.8 &8.5  &34.2    &62.5 &10.6 & 26.9  &53.7 &10.2 & 36.1  &55.2 &16.8 & 28.0  &0.73 \\
 & DTR vs. StylizedDU   &49.2 &16.8 & 34.0   &55.8 &24.9 & 19.3  &50.9 &22.4 & 26.7  &38.7 &25.1 & 36.2  &0.66 \\
 & DTR vs. StyleDGPT    &56.7 &21.6 & 21.7    &51.6 &27.4 & 21.0  &54.0 &22.6 & 23.4  &52.6 &22.9 & 24.5  &0.64 \\
\hline
\multirowcell{3}{Polite} 
& DTR vs. StyleFusion   &58.2 &12.6 & 29.2    &56.5 &8.9  & 34.6  &58.3 &11.6 & 30.1  &50.7 &23.1 & 26.2  &0.68 \\
 & DTR vs. StylizedDU   &54.6 &17.1 & 28.3    &48.0 &23.8 & 28.2  &48.2 &25.1 & 26.7  &46.0 &28.3 & 25.7  &0.70 \\
 & DTR vs. StyleDGPT    &49.8 &19.3 & 30.9    &46.5 &28.1 & 25.4  &45.6 &27.3 & 27.1  &53.5 &21.1 & 25.4  &0.65 \\
\hline
\end{tabular}
\caption{\label{font-table}  Manual evaluation results.  W, L, and T refer to Win, Lose, and Tie, respectively. The ratios are calculated
by combining labels from the three annotators.
}
\label{res:human_all}
\end{table*}

\subsection{Replace Rate $P_r$} \label{appendix:replace}
\begin{figure}[ht]
\centering
     \includegraphics[width=0.5\textwidth, scale=0.85, trim=15 6 0 35,clip]{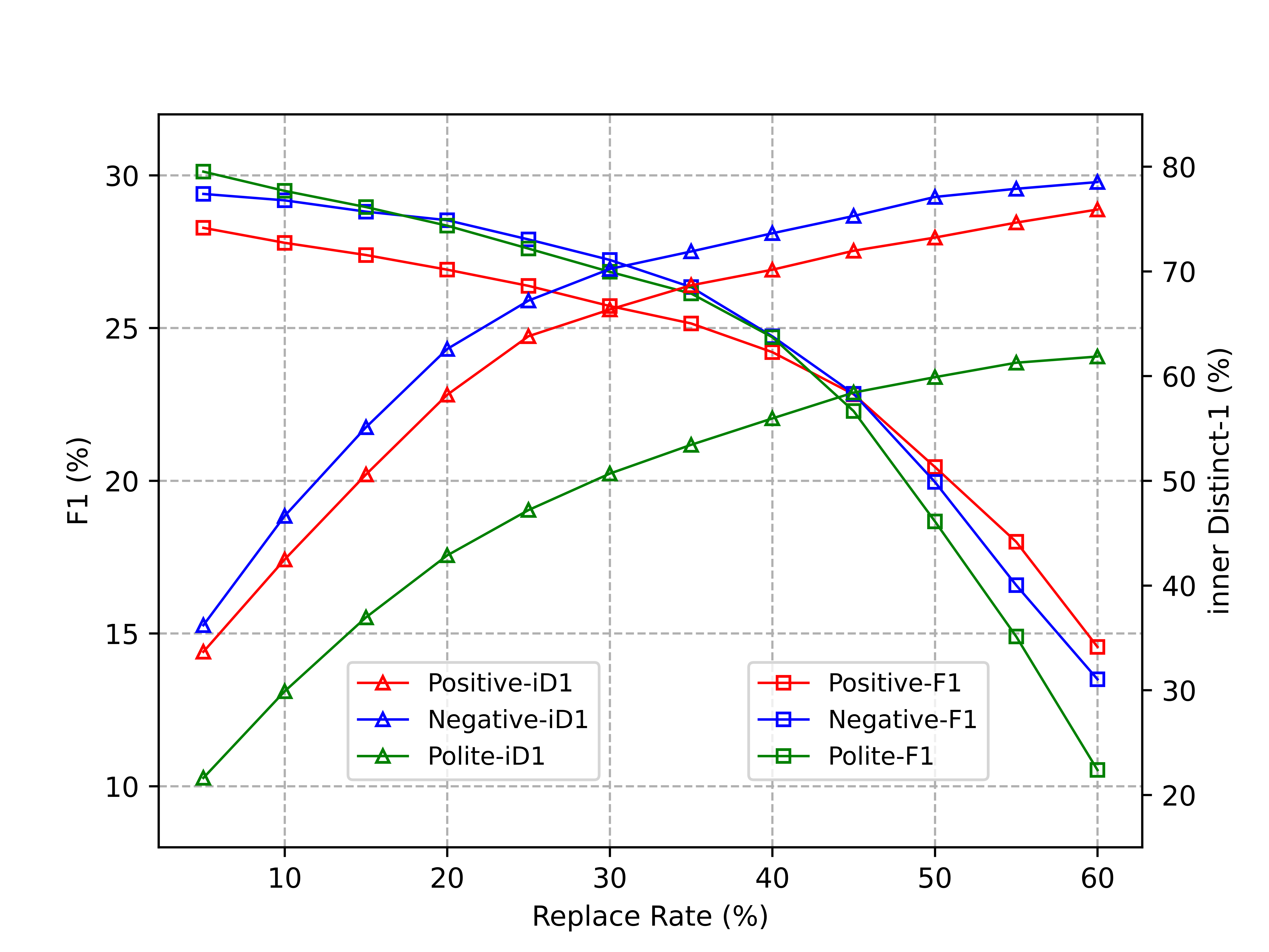}
     \caption{F1 and Inner Distinct with different replace rate in Topical Chat.}
     \label{fig:mask_topic}
\end{figure}

As shown in Figure \ref{fig:mask_topic}, we present the F1 and Inner Distinct with different replace rate in Topical Chat.

\subsection{Statistics of frequent style words} \label{appendix:stawords}
\begin{figure}[ht]
\centering
     \includegraphics[width=0.5\textwidth, scale=1, trim=260 340 415 165,clip]{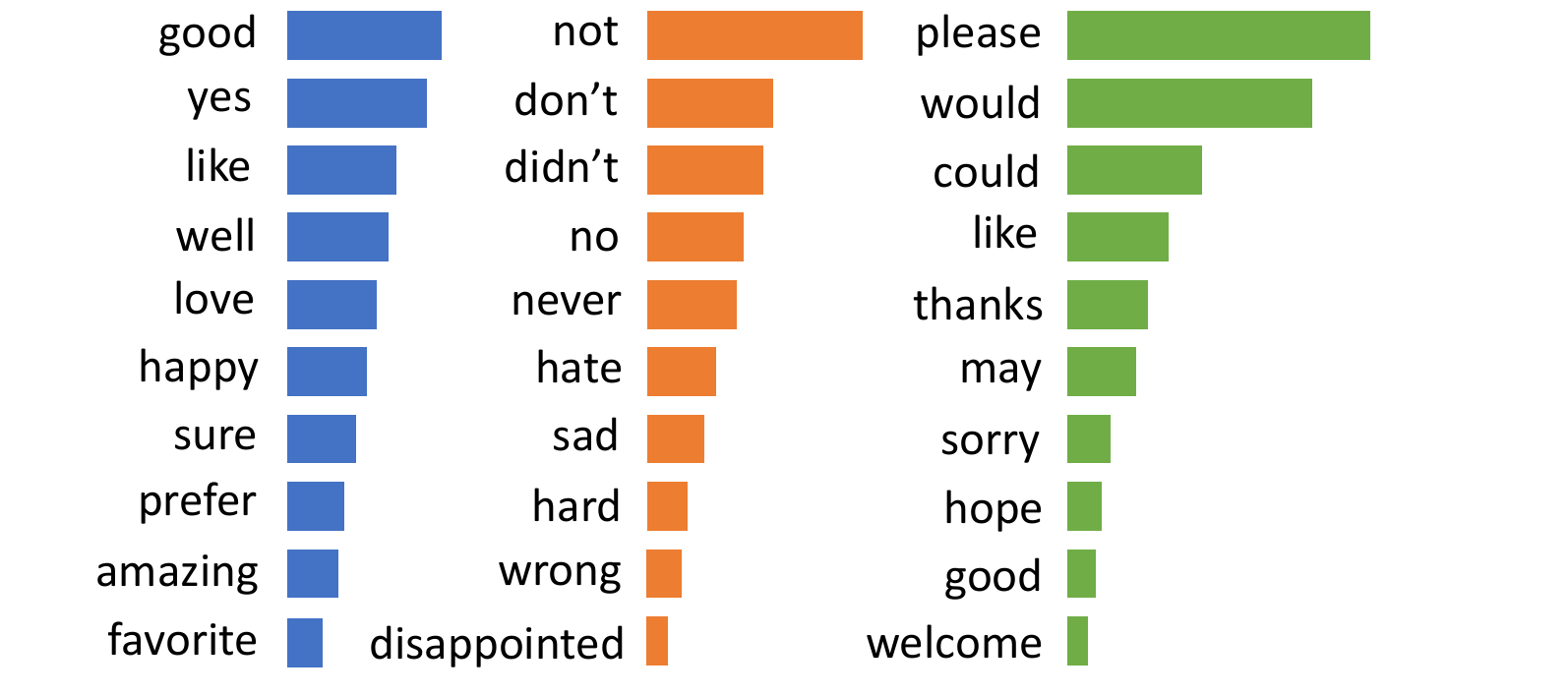}
     \caption{Statistics of frequent new generated words in Positive, Negative, and Polite.}
     \label{fig:keywords}
\end{figure}

As shown in Figure \ref{fig:keywords}, we present the visualization of
the style tokens in various style corpus found by the initiated style decoupler.

\subsection{F1 Drop  $P_r$} \label{appendix:f1}
\begin{figure}[ht]
\centering
     \includegraphics[width=0.5\textwidth, scale=0.85, trim=15 6 0 35,clip]{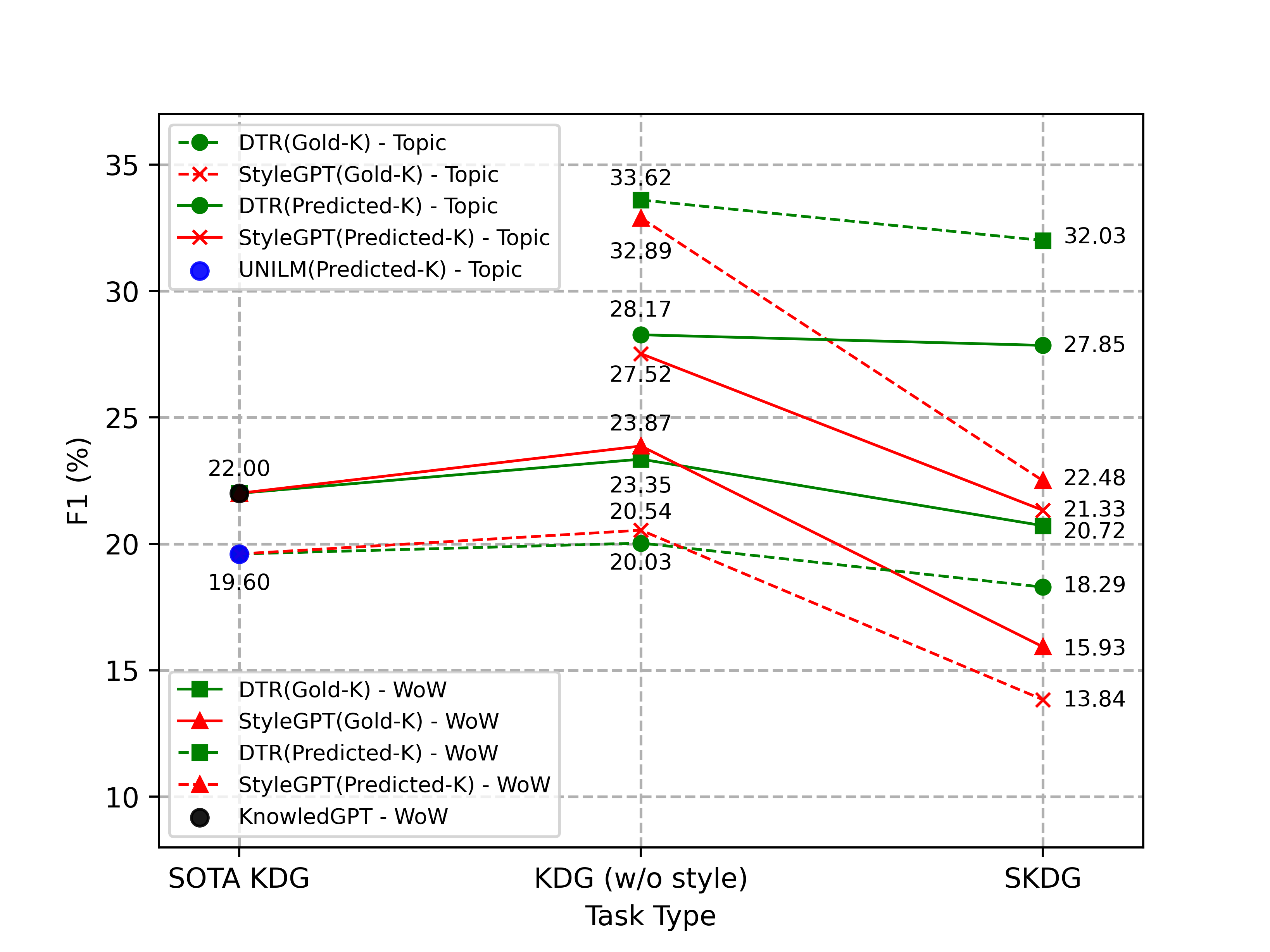}
     \caption{F1 of DTR, StyleDGPT, and SOTA KDG models in different task mode of negative sentiment. Gold-K represents we use ground truth knowledge as input, and Predicted-K represents we use a knowledge selection model to predict top-1 knowledge sentence as input.}
     \label{fig:dropnegative}
\end{figure}

\begin{figure}[ht]
\centering
     \includegraphics[width=0.5\textwidth, scale=0.85, trim=15 6 0 35,clip]{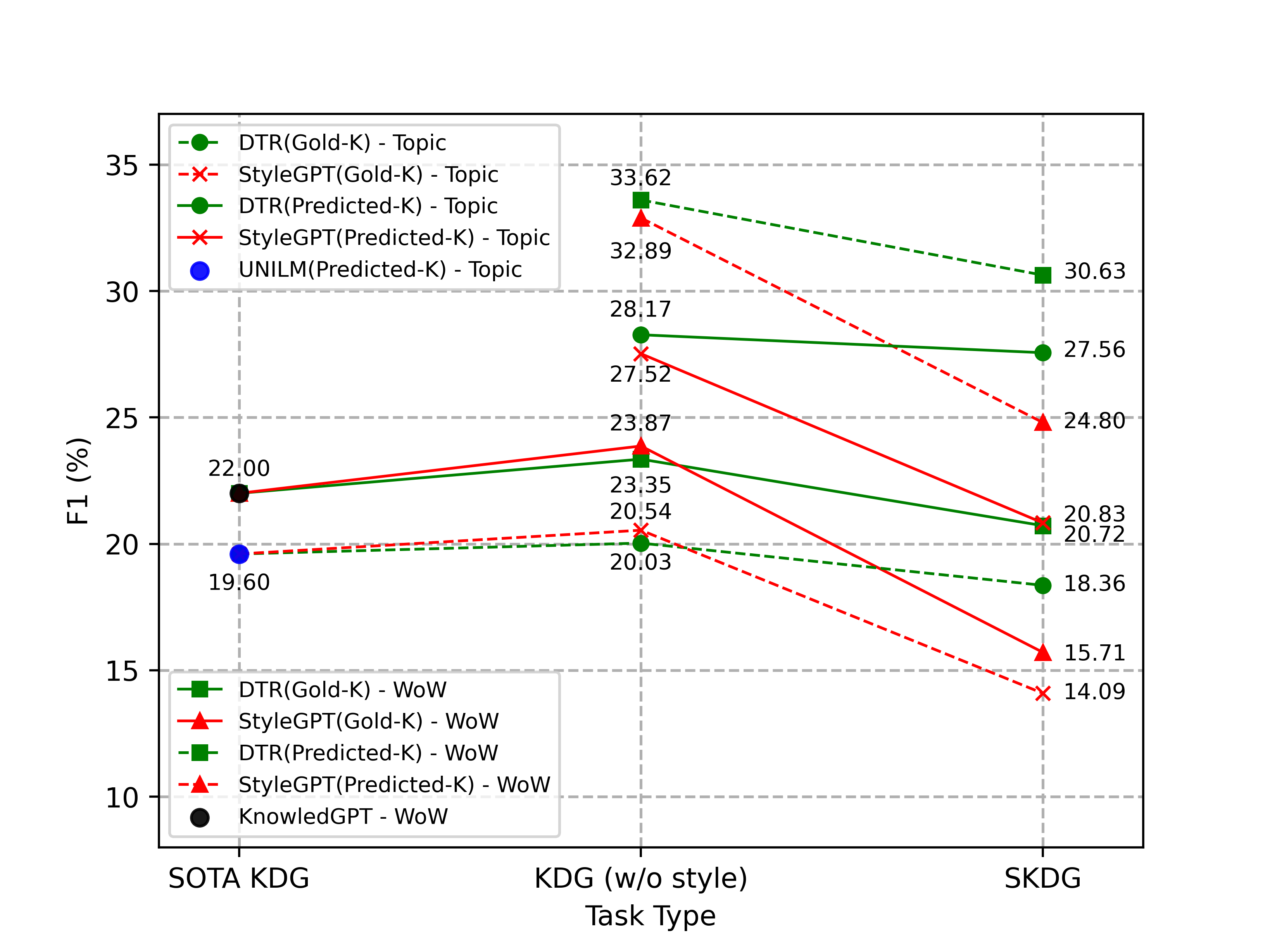}
     \caption{F1 of DTR, StyleDGPT, and SOTA KDG models in different task mode of polite style. Gold-K represents we use ground truth knowledge as input, and Predicted-K represents we use a knowledge selection model to predict top-1 knowledge sentence as input.}
     \label{fig:droppolitee}
\end{figure}

As shown in Figure \ref{fig:dropnegative} and \ref{fig:droppolitee} , we present F1 of DTR, StyleDGPT, and SOTA KDG models in different task mode of negative sentiment and polite style.



\end{document}